# Efficient and generalizable prediction of molecular alterations in multiple cancer cohorts using H&E whole slide images

[1]Kshitij Ingale, [1]Sun Hae Hong, [1]Qiyuan Hu, [2]Renyu Zhang, [1]Bo Osinski, [2]Mina Khoshdeli, [2]Josh Och, [1]Kunal Nagpal, [2]Martin C. Stumpe, [1]Rohan P. Joshi

**Author Affiliations**

1. Tempus AI, Inc. Chicago, IL
2. Work done while at Tempus AI, Inc. Chicago, IL

# Abstract

Molecular testing of tumor samples for targetable biomarkers is restricted by a lack of standardization, turnaround-time, cost, and tissue availability across cancer types. Additionally, targetable alterations of low prevalence may not be tested in routine workflows. Algorithms that predict DNA alterations from routinely generated hematoxylin and eosin (H&E)-stained images could prioritize samples for confirmatory molecular testing. Costs and the necessity of a large number of samples containing mutations limit approaches that train individual algorithms for each alteration. In this work, models were trained for simultaneous prediction of multiple DNA alterations from H&E images using a multi-task approach. Compared to biomarker-specific models, this approach performed better on average, with pronounced gains for rare mutations. The models reasonably generalized to independent temporal-holdout, externally-stained, and multi-site TCGA test sets. Additionally, whole slide image embeddings derived using multi-task models demonstrated strong performance in downstream tasks that were not a part of training. Overall, this is a promising approach to develop clinically useful algorithms that provide multiple actionable predictions from a single slide.

# Introduction

Molecular alterations have predictive, prognostic, and diagnostic value across various cancer types [1–3]. Due to factors including tissue availability, cost, turnaround time, and low prevalence of actionable biomarkers, molecular testing is not always ordered for patients who may benefit from it. In contrast, cancer diagnosis nearly always involves a tissue biopsy with hematoxylin and eosin (H&E)-stained histopathology slides routinely acquired in the clinical workflow. H&E stained slides are increasingly being digitized as whole slide images (WSIs) to assist pathology workflows and for archival purposes [4]. Therefore, predicting molecular alterations from H&E WSIs could identify patients who are likely to have actionable genetic mutations. These predicted positive patients may benefit from confirmatory testing to evaluate their eligibility for targeted therapies and other downstream implications. In cases of limited tissue availability, prediction of alterations could also assist in prioritizing tissue for downstream assays, such as for immunohistochemistry stains, rapid small panel molecular tests, or time-consuming but comprehensive next-generation sequencing (NGS).

The application of machine learning to inferring biomarkers from routine H&E-stained WSIs has the potential to rapidly screen patients for molecular alterations and prioritize patients for confirmatory testing [5]. However, models in these studies were trained for single biomarker prediction, resulting in numerous models to cover all alterations of interest. This approach not only is more computationally expensive, but it also does not leverage the correlations between different targets to improve model performance via joint learning. It can also be challenging to build single target models for low-prevalence actionable biomarkers as deep learning algorithms learn from large datasets [6].

In other domains, multi-task learning has demonstrated the ability to leverage information learned by one task to benefit the training of other related tasks [7–9]. However, this approach has only been studied to a limited extent in the prediction of biomarkers from histopathology images. To the best of our knowledge, detailed comparisons of the performance of multi-task and single-task approaches for biomarker prediction from H&E images have not been performed. In particular, which biomarkers benefit from a multi-task approach and how well the model approaches generalize have not been examined. Indeed, the generalizability of histopathology machine learning algorithms across multi-site staining and scanning characteristics remains a significant challenge in the field, and algorithm validation across external pre-analytic characteristics remains important for determining clinical utility [10,11].

In this study, we curated a large library of paired imaging and molecular data across multiple cancer types and many DNA molecular alterations. We then developed models to predict the broad range of molecular alterations from H&E WSIs using attention-based multiple instance learning using either a per-cancer multi-task or per-cancer single-task approach. We rigorously compared multi-task models to single-task models across a multitude of targets and demonstrated that multi-task models were not only significantly less computationally expensive, but also significantly enhanced the performance on targets with low prevalence. We further evaluated the generalizability of these models to slides stained at external labs and slides sourced from The Cancer Genome Atlas (TCGA). Finally, we demonstrated that the embeddings produced by these models can be used to train models for tasks that the original network was not trained on, suggesting the network is learning informative, generalizable representations of the histology that have the potential to be efficiently utilized in other applications. The multi-target histogenomics models presented in this work provide an efficient and generalizable method for simultaneously predicting actionable molecular alterations using H&E WSIs and demonstrate potential as a pre-screen to improve molecular testing of patients for personalized cancer treatment.

# Methods

## Cohort curation

A retrospective dataset of digitized H&E images was curated consisting of 12,883 colorectal cancer, 8,805 breast cancer, 2,900 endometrial cancer, and 2,873 bladder cancer patient whole slide images (Supplemental Table 1-4). Since some tissue extraction methods such as bone marrow core biopsy, fine needle aspiration, fluid aspiration, skin punch, or venipuncture can affect histology tissue architecture integrity, these cases were excluded from the cohort. For patients with multiple slides, a single slide was sampled. Most of the slides were prepared at the in-house lab (Supplemental Table 1-4), while some were stained at the ordering institution and shipped to the in-house lab, where the slide was digitized. The slides were digitized using a Philips UFS scanner (Philips, Amsterdam, Netherlands) or Aperio GT450 scanner (Leica Biosystems, Wetzlar, Germany). DNA alterations for these tissue samples were detected through a 648-gene DNA panel (Tempus AI, Inc., Chicago, IL, USA). MSI status for internal cases was defined using a clinically validated internal pipeline [12].

DNA alterations were selected for analysis if at least 200 patients in each cohort had pathogenic or likely pathogenic mutations. In the endometrial cancer cohort, *POLE* (DNA Polymerase Epsilon, Catalytic Subunit) alterations are important for molecular subtyping, so *POLE* alteration was also added despite there being fewer than 200 patients with *POLE* mutations in the cohort.

## Data splitting

Three validation sets were constructed to evaluate model performance (Figure 1). First, slides prepared at external institutions were held out as an externally stained holdout test set to evaluate model generalization to diverse histopathological staining characteristics. Next, 20% of the most recent samples received for sequencing were held out to form a temporal holdout test set. Finally, to further evaluate model generalization, models were also evaluated on the publicly available multi-site dataset from The Cancer Genome Atlas (TCGA). The diagnostic slides were downloaded using the Genomic Data Commons (GDC) tool [13] and the first diagnostic slide was chosen for each case. DNA alteration labels were determined using OncoKB annotations [14] identified mutations as oncogenic or likely oncogenic. Microsatellite instability (MSI) labels were determined using MSISensor scores [15] with positive cases identified as scores greater than 10. Clinical data for TCGA cohorts corresponding to the TCGA Pan-Cancer analysis project [16] were obtained through the cbioportal website [17]. TCGA slides missing ICC profile information were excluded from the analysis.

The remaining internally prepared cases were used for developing models (Supplemental Table 1-4). The model development data were split into five folds for cross-validation, where three folds were used for training, one fold was used for model selection, and one fold for testing. An iterative process was used to stratify the cross-validation splits based on multiple labels of interest, digitization scanner, tissue site, and procedure type [18].

## Dataset preprocessing

Regions containing pertinent tissue and excluding artifacts and pathologist marker annotations were identified using a previously developed U-Net model [19]. Subsequently, tiles of size 224 x 224 pixels at 10x magnification were obtained from tissue regions with at least 80% of the tile containing tissue (Figure 1). Images were transformed to standard RGB space by applying ICC profiles to aid in generalization [20] and normalized before fed into the model.

## Model architecture and training

For each cancer cohort, a multi-task learning model based on attention-based multiple instance learning mechanism [21] was trained to simultaneously predict multiple DNA alterations from H&E images. This model used a standard pretrained ResNet18 [22] feature extractor, followed by an attention module to identify tiles with high diagnostic relevance and a classifier module aggregating all information to predict all DNA alteration targets (Figure 1). This architecture was compared against individual biomarker models optimized to predict single targets. The model processes a randomly sampled set of tiles to generate features and attention scores and aggregates them to yield a slide-level prediction.

The models were trained end-to-end using Adam optimizer [23] with a learning rate of 1 x $10^{-5}$ and weight decay of 1 x $10^{-4}$. For the multi-task model, the average weighted cross-entropy across all targets was optimized, while for individual biomarker models, weighted cross-entropy loss for each target was optimized. The weighing scheme was based on the reversed prevalence of targets; for example, *FGFR3* alteration prevalence in bladder cancer cohort was around 13%, resulting in weights of 0.13 applied to the negative class and 0.87 to the positive class. Training was performed across 4 NVIDIA T4 GPUs (Nvidia, Santa Clara, CA, USA). During training, a set of 100 tiles per slide was fed to the model and the effective batch size was 16. During inference, 1000 tiles per slide were used to provide additional context for inference, and the batch size was 1. Tiles were sampled from each slide without replacement, with all tiles being used if the number of tiles was less than the bag size (100 during training and 1000 during inference). Data augmentation was performed to improve robustness through spatial jitter, color jitter, random horizontal and vertical flip, and random rotations by 90 degrees.

Model training, inference, and analyses were performed using torch v1.13.1, torchvision v0.14.1, scikit-learn v1.4.0, statsmodels v0.14.1, scipy v1.12.0 libraries in Python language v3.10.13. Cohort tables were generated using the gtsummary v1.7.2 package in R language v4.3.1.

## Model performance evaluation

Cohort-specific multi-task models were trained to simultaneously predict 58 alterations in colorectal cancer, 53 alterations in breast cancer, 24 alterations in bladder cancer and 17 alterations in endometrial cancer cohorts. To quantify the effect of multi-task training relative to single-task training, single-task models were trained for the same alterations in endometrial cancer and 15 randomly sampled targets in colorectal cancer reflecting the entire range of prevalences. Area under the receiver operating characteristic curve (ROC-AUC) was calculated for these models in cross-validation folds. The probabilities from cross-validation fold models were averaged to obtain mean ensemble scores during evaluation.

## Attention scores and embeddings analysis

Tiles were labeled using a custom histology image classifier [24] as tumor, stroma, epithelium, necrosis, immune, and other for 1989 colorectal cancer and 1333 breast cancer test set slides. High attention tiles were identified in each slide as 10% of the tiles with the highest attention scores. The proportions of tile classes were compared among high-attention tiles and all tiles. Wilcoxon signed-rank test was performed to assess whether tumor tiles were overrepresented in high attention tiles relative to all tiles.

For a subset of externally stained test set images, tumor annotations were performed by board-certified pathologists at 10x magnification. Binary tumor labels were obtained for all the tiles with positive class comprising more than 50% tumor region. ROC-AUC was calculated with attention probabilities to compare binary tumor classes. This test set was resampled to generate 10,000 bootstrap samples and variability was measured by 95th percentile interval of the ROC-AUC values on these bootstrap samples.

Because shared learning among different tasks can allow models to learn useful feature representation for downstream correlated tasks [25], embeddings from the multi-task model were extracted from the classifier layer for each of the four cancer types. Logistic regression models were trained on embeddings to predict binary cancer grade (high vs low) and whether a slide is from the primary tissue site or not. ROC-AUC was calculated on a test set to evaluate the logistic regression model.

## Statistical analysis

The mean cross-validation ROC-AUC scores from multi-task and individual biomarker models were compared using a paired t-test. Paired t-test was performed to compare ROC-AUC between the two model types on temporal, externally stained, and TCGA sets. Pearson correlation test was used to study the relationship between target prevalence and the increase in ROC-AUC for multi-task models over individual models. A paired t-test was also performed to compare the drop in ROC-AUC from the cross-validation set to the holdout test sets between the two model types to evaluate their generalizability. The ratio of tile classes in high-attention tiles was compared to that in all tiles using a one-tailed Wilcoxon signed-rank test. The significance level for hypothesis testing was set to 0.05 in these experiments.

# Results

## Multi-task models outperform individual biomarker models especially in rare targets

A multi-task learning approach was compared with single-task biomarker-specific models in colorectal cancer and endometrial cancer cohorts (15 targets in colorectal cancer, 17 targets in endometrial cancer, see Methods and Figure 1). Multi-task models yielded significantly higher ROC-AUCs in colorectal cancer, with a mean ROC-AUC of 0.72 ± 0.10 s.d. vs 0.67 ± 0.13 s.d. ($p$ = 0.001, one-tailed paired t-test, Figure 2a). Similar results were found in endometrial cancer, with a mean ROC-AUC of 0.72 ± 0.09 s.d. vs 0.68 ± 0.10 s.d. ($p$ < 0.0001, one-tailed paired t-test, Figure 2b).

Because building biomarker-specific models for rare targets is challenging due to small sample size, multi-task models can be helpful. Indeed, a significant correlation between gain in performance and prevalence was found in colorectal cancer (Pearson r = -0.84, $p$ < 0.0001, Figure 2c) and endometrial cancer (Pearson r = -0.79, $p$ < 0.0001, Figure 2d) cohorts.

Compared to biomarker-specific models, multi-task model training time and compute cost were decreased by orders of magnitude (58-fold for colorectal, 53-fold for breast, 24-fold for bladder and 17-fold for endometrial cancer).

# Multi-task models yield higher ROC-AUC than individual biomarker models on holdout test sets

In order to account for the wide variety of data encountered by machine learning models in real-world deployment, model robustness was tested for time variability, lab-specific stain differences, and institution batch effects. Robustness of the model to time variability was evaluated using the temporal holdout test set. The mean temporal test set ROC-AUC achieved by multi-task models was significantly higher than individual biomarker models for targets in colorectal cancer (0.74 ± 0.10 s.d. vs. 0.70 ± 0.13 s.d., $p$ = 0.004 for one-tailed t-test, N=15 targets, Figure 3a) and endometrial cancer (0.73 ± 0.08 s.d. vs 0.70 ± 0.10 s.d., $p$ = 0.03 for one-tailed t-test, N=16 targets, Figure 3b). Slides prepared at external labs may have color differences owing to differences in lab staining protocols. In an external test set, multi-task models also significantly outperformed individual biomarker models in colorectal cancer (0.72 ± 0.09 s.d. vs. 0.68 ± 0.11 s.d., $p$ = 0.005 for one-tailed t-test, N=15 targets, Figure 3c) and endometrial cancer (0.72 ± 0.11 s.d. vs 0.68 ± 0.12 s.d., $p$ < 0.0001 for one-tailed t-test, N=15 targets, Figure 3d). Data acquired at other institutions may additionally have both different patient characteristics and image staining qualities, as found in TCGA. On TCGA test sets, multi-task models mean ROC-AUC trended higher than individual biomarker models in colorectal cancer (0.74 ± 0.08 s.d. vs. 0.70 ± 0.11 s.d., $p$ = 0.06 for one-tailed t-test, N=8 targets, Figure 3e) and was significantly higher than individual biomarker models in endometrial cancer (0.67 ± 0.08 s.d. vs 0.65 ± 0.09 s.d., $p$ = 0.003 for one-tailed t-test, N=17 targets, Figure 3f).

# Multi-task models can screen biomarkers in Breast cancer and Bladder cancer cohorts

Given the superior performance of multi-task models in colorectal and endometrial cancer cohorts, multi-task models were also trained for breast and bladder cohorts. Additional targets were also examined within the multi-task training framework for the colorectal cancer cohort. The average ROC-AUC across all targets in cross-validation was 0.73 ± 0.10 s.d. in colorectal cancer (N=58), 0.72 ± 0.09 s.d. in endometrial cancer (N=17), 0.64 ± 0.07 s.d., in breast cancer (N=53), and 0.60 ± 0.07 s.d. in bladder cancer (N=24, Figure 4a-d). For the temporal holdout test set, the average ROC-AUC was 0.75 ± 0.11 s.d. (N=57) in colorectal cancer, 0.73 ± 0.08 s.d. (N=16) in endometrial cancer, 0.65 ± 0.08 s.d. (N=51) in breast cancer, 0.64 ± 0.08 s.d. (N=24) in bladder cancer (Figure 5a-d). The average test set ROC-AUC on externally stained slides was 0.72 ± 0.11 s.d. (N=58) in colorectal cancer, 0.72 ± 0.11 s.d. (N=15) in endometrial cancer, 0.65 ± 0.08 s.d. (N=53) in breast cancer, 0.64 ± 0.08 s.d. (N=24) in bladder cancer (Figure 5a-d). Across the multi-site TCGA test set, mean ROC-AUC was 0.68 ± 0.10 s.d. (N=15) in colorectal cancer, 0.67 ± 0.08 s.d. (N=17) in endometrial cancer, 0.60 ± 0.09 s.d. (N=18) in breast cancer, 0.60 ± 0.11 s.d. (N=15) in bladder cancer (Figure 5a-d). Additionally, compared to in-house generated temporal holdout test set, performance on externally stained slides was comparable and was slightly lower on TCGA test set (Supplemental Figure S1 and roc curves for a subset of targets in Supplemental Figure S2, see supplemental information for comparison method).

# Multi-task models assign high attention to tumor tiles

The attention scores assigned to each tile by the model determine the tile's influence on the final prediction. A custom histology classifier model assigned labels to the tiles, and the histology class distribution was analyzed for the high-attention tiles. Tumor regions constituted a significantly larger fraction of high attention tiles than all the tiles used for making predictions in colorectal cancer slides (all tiles mean fraction 0.34 ± 0.21 s.d. vs high attention tiles mean fraction 0.80 ± 0.24 s.d., $p < 0.0001$ by one-tailed Wilcoxon test) and breast cancer slides (all tiles mean fraction 0.43 ± 0.27 s.d. vs high attention tiles mean fraction 0.83 ± 0.26 s.d., $p < 0.0001$ by one-tailed Wilcoxon test) (Figure 6a-b).

The model attention scores were also compared against board-certified pathologists' tumor annotations. An ROC curve was used to analyze the relationship between tile attention scores and corresponding tumor annotation labels. The score was indicative of tumor region with an ROC-AUC of 0.81 in colorectal cancer (95th percentile interval: 0.80 to 0.82) and 0.83 in breast cancer (95th percentile interval: 0.82 to 0.84, Figure 7a-b). A qualitative comparison also indicates a strong overlap between attention probability heatmap and tumor annotations (Figure 7c-d).

## Multi-task models learn a versatile feature representation

The shared learning from training on multiple tasks could help models learn a feature representation that is predictive for other correlated tasks that were not part of training. The attention-based neural network trained for the DNA target tasks above creates embeddings that are representative of the slide. Logistic regression models trained to predict cancer grade using these embeddings were able to achieve ROC-AUC of 0.86 (95th percentile interval: 0.81 to 0.90) in colorectal cancer, 0.83 (95th percentile interval: 0.74 to 0.92) in endometrial cancer, 0.81 (95th percentile interval: 0.76 to 0.86) in breast cancer, and 0.90 (95th percentile interval: 0.84 to 0.96) in bladder cancer on a withheld test set (Figure 8a-d). Similarly, models trained to predict whether the slide is from the primary tissue site or not achieved holdout test ROC-AUC of 0.94 (95th percentile interval: 0.92 to 0.95) in colorectal cancer, 0.87 (95th percentile interval:

0.81 to 0.91) in endometrial cancer, 0.93 (95th percentile interval: 0.91 to 0.95) in breast cancer, and 0.81 (95th percentile interval: 0.74 to 0.88) in bladder cancer (Figure 8e-h).

## Discussion

Identification of molecular alterations in tumor DNA is a critical step in cancer treatment [26]. Advances in precision medicine allow the identification of targetable biomarkers, while AI tools can be used to predict individual molecular alterations from H&E WSIs. These tools may help prioritize patients who would benefit from subsequent confirmatory testing [27,28]. However, the development of separate AI tools for individual mutations is a resource-intensive endeavor that does not take advantage of the shared biology that could be present between mutations in different genes. In this work, we used a large cohort of real-world samples to demonstrate that DNA alterations could be effectively learned in a multi-task manner from histopathology images. Multi-task training was especially beneficial in identifying low-prevalence molecular alterations. These models were generalizable to a temporally-independent cohort, an externally-stained cohort, and a multi-institutional cohort consisting of different sequencing protocols, staining characteristics and scanning devices.

Panel-based NGS sequencing approaches can confirm that mutations are present in a wide variety of genes using a single assay. For example, non-small cell lung cancer can present with low-frequency but targetable alterations in *EGFR*, *KRAS*, *MET*, *HER2*, *ROS*, *ALK*, and other genes and is now the standard of care. Rates of testing are affected by cost and patient tumor tissue volume required to perform NGS and other additional specialized immunohistochemistry tests. An H&E multi-task model approach could provide a similar gene panel to NGS that screens multiple targets with a single H&E image, thus prioritizing patient tissue and health care resources towards the appropriate confirmatory test. In the four cancer types investigated, our multi-task approach demonstrated the ability to create these "H&E panels" with good performance for both common and rare mutations.

While the rarest mutations are the most likely to benefit from an H&E screening-based approach, building models can be challenging as deep learning approaches typically require a large number of positive samples. These rare mutations, however, may have biological processes shared with more common mutations that result in similar H&E morphological characteristics. For example, in endometrial cancer, both MSI and *POLE* mutations may result in similar immune cell infiltration of the tumor. Studies have demonstrated the benefit of multi-task training on label-image pairs that are correlated [29]. Consistent with this, the multi-task model approach demonstrated a pronounced gain in performance for rare targets compared to biomarker-specific models. This is a highly beneficial property, given that screening in medicine is most important when prevalence is low and patient impact is high.

In real-world applications, models using H&E images can encounter a wide range of image characteristics due to different lab staining protocols, and scanning devices. The multi-task models reasonably generalized to a temporally held out test set and an externally stained test set, indicating robustness to temporal drift in data characteristics and staining patterns at external labs. Data from multiple institutions like TCGA, can have different patient characteristics as well. Multi-task models suffered a minor decrease in performance on the TCGA test sets. Multi-task models were also able to achieve better performance than single-task models on average across these test sets than individual biomarker models.

Model generalizability is connected to the ability to capture true biological signals present in H&E images related to genetic alterations. Consistent with known biology, tumor regions were overrepresented among high-attention tiles of the model and attention probabilities from the model were also able to identify tumor and non-tumor regions annotated by board-certified pathologists. Also, qualitative comparison of tumor region annotations showed high overlap with attention heatmaps. These findings suggest that the model used biologically relevant regions for making predictions as molecular alterations usually result in morphological changes in tumor areas.

There are several limitations to this study. While rare mutations benefited from multi-task training, high-prevalence alterations did not. Although future work may address this gap, screening for low-prevalence alterations has a higher clinical value, and developing algorithms that can detect low-prevalence alterations has proven to be more challenging. Moreover, the models presented here generalized well to retrospectively-acquired data. For real-world clinical deployment, the models must be further validated through rigorous prospective studies to establish clinical value.

While this work demonstrated value in four cancer types, future work could adopt a modeling approach unifying across all cancer types and use a single model, multiple-cancer multiple-target approach. Additionally, we demonstrated in a preliminary analysis that the multi-task model embeddings were able to predict cancer grade and primary tissue site for bladder, breast, colorectal, and endometrial cancer cohorts, which indicates that the robust feature representation learnt by the model can be leveraged for other impactful tasks that were not part of training, a property similar to that of foundation models. The approach taken here should be compared to a self-supervised foundational model on a more comprehensive set of tasks in future work.

Ultimately, we believe that the work presented here has high value for the creation of "H&E panel" AI algorithms that can take an H&E slide, screen for many actionable molecular alterations, and provide results to an oncologist to optimize patient sample testing.

# Data and code availability

# Acknowledgements

We thank Dana DeSantis and Adam Hockenberry for reviewing the manuscript. We thank Yoni Muller and Robert Montgomery for providing operational support.

# References

1. Brandner, S. & von Deimling, A. Diagnostic, prognostic and predictive relevance of molecular markers in gliomas. *Neuropathol. Appl. Neurobiol.* **41**, 694–720 (2015).

2. Koncina, E., Haan, S., Rauh, S. & Letellier, E. Prognostic and Predictive Molecular Biomarkers for Colorectal Cancer: Updates and Challenges. *Cancers* **12**, 319 (2020).

3. Gurel, B. *et al.* Molecular Alterations in Prostate Cancer as Diagnostic, Prognostic, and Therapeutic Targets. *Adv. Anat. Pathol.* **15**, 319 (2008).

4. Pantanowitz, L. Digital images and the future of digital pathology. *J. Pathol. Inform.* **1**, 15 (2010).

5. Juan Ramon, A. *et al.* Development and deployment of a histopathology-based deep learning algorithm for patient prescreening in a clinical trial. *Nat. Commun.* **15**, 4690 (2024).

6. Lee, J. *et al.* Deep learning for rare disease: A scoping review. *J. Biomed. Inform.* **135**, 104227 (2022).

7. Ruder, S. An Overview of Multi-Task Learning in Deep Neural Networks. Preprint at https://doi.org/10.48550/arXiv.1706.05098 (2017).

8. Zhang, Y. & Yang, Q. A Survey on Multi-Task Learning. *IEEE Trans. Knowl. Data Eng.* **34**, 5586–5609 (2022).

9. Collobert, R. & Weston, J. A unified architecture for natural language processing: deep neural networks with multitask learning. in *Proceedings of the 25th international conference on Machine learning* 160–167 (Association for Computing Machinery, New York, NY, USA, 2008). doi:10.1145/1390156.1390177.

10. Kleppe, A. *et al.* Designing deep learning studies in cancer diagnostics. *Nat. Rev. Cancer* **21**, 199–211 (2021).

11. Howard, F. M. *et al.* The impact of site-specific digital histology signatures on deep learning model accuracy and bias. *Nat. Commun.* **12**, 4423 (2021).

12. Beaubier, N. *et al.* Clinical validation of the tempus xT next-generation targeted oncology sequencing assay. *Oncotarget* **10**, 2384–2396 (2019).

13. Grossman Robert L. *et al.* Toward a Shared Vision for Cancer Genomic Data. *N. Engl. J. Med.* **375**, 1109–1112 (2016).

14. Chakravarty, D. *et al.* OncoKB: A Precision Oncology Knowledge Base. *JCO Precis. Oncol.* **1**, PO.17.00011 (2017).

15. Niu, B. *et al.* MSIsensor: microsatellite instability detection using paired tumor-normal sequence data. *Bioinformatics* **30**, 1015–1016 (2014).

16. Weinstein, J. N. *et al.* The Cancer Genome Atlas Pan-Cancer analysis project. *Nat. Genet.* **45**, 1113–1120 (2013).

17. Cerami, E. *et al.* The cBio cancer genomics portal: an open platform for exploring multidimensional cancer genomics data. *Cancer Discov.* **2**, 401–404 (2012).

18. Sechidis, K., Tsoumakas, G. & Vlahavas, I. On the Stratification of Multi-label Data. in *Machine Learning and Knowledge Discovery in Databases* (eds. Gunopulos, D., Hofmann, T., Malerba, D. & Vazirgiannis, M.) 145–158 (Springer, Berlin, Heidelberg, 2011). doi:10.1007/978-3-642-23808-6_10.

19. Ronneberger, O., Fischer, P. & Brox, T. U-Net: Convolutional Networks for Biomedical Image Segmentation. in *Medical Image Computing and Computer-Assisted Intervention – MICCAI 2015* (eds. Navab, N., Hornegger, J., Wells, W. M. & Frangi, A. F.) 234–241 (Springer International Publishing, Cham, 2015). doi:10.1007/978-3-319-24574-4_28.

20. Ingale, K., Joshi, R. P., Ho, I. Y., BenTaieb, A. & Stumpe, M. C. Effects of Color Calibration via ICC Profile on Inter-scanner Generalization of AI Models IN USCAP 2022 Abstracts: Informatics (977-1017). *Mod. Pathol.* **35 Suppl 2**, 1163–1210 (2022).

21. Ilse, M., Tomczak, J. & Welling, M. Attention-based Deep Multiple Instance Learning. in *Proceedings of the 35th International Conference on Machine Learning* 2127–2136 (PMLR, 2018).

22. He, K., Zhang, X., Ren, S. & Sun, J. Deep Residual Learning for Image Recognition. in *2016 IEEE Conference on Computer Vision and Pattern Recognition (CVPR)* 770–778 (2016). doi:10.1109/CVPR.2016.90.

23. Kingma, D. P. & Ba, J. Adam: A Method for Stochastic Optimization. Preprint at https://doi.org/10.48550/arXiv.1412.6980 (2017).

24. Sha, L. *et al.* Multi-Field-of-View Deep Learning Model Predicts Nonsmall Cell Lung Cancer Programmed Death-Ligand 1 Status from Whole-Slide Hematoxylin and Eosin Images. *J. Pathol. Inform.* **10**, 24 (2019).

25. Maurer, A., Pontil, M. & Romera-Paredes, B. The Benefit of Multitask Representation Learning. *J. Mach. Learn. Res.* **17**, 1–32 (2016).

26. Coyle, K. M., Boudreau, J. E. & Marcato, P. Genetic Mutations and Epigenetic Modifications: Driving Cancer and Informing Precision Medicine. *BioMed Res. Int.* **2017**, 9620870 (2017).

27. Ingale, K. *et al.* Prediction of MET Overexpression in Lung Adenocarcinoma from Hematoxylin and Eosin Images. *Am. J. Pathol.* **194**, 1020–1032 (2024).

28. Hu, Q. *et al.* Development and validation of a deep learning-based microsatellite instability predictor from prostate cancer whole-slide images. *NPJ Precis. Oncol.* **8**, 88 (2024).

29. Nahhas, O. S. M. E. *et al.* Joint multi-task learning improves weakly-supervised biomarker prediction in computational pathology. Preprint at https://doi.org/10.48550/arXiv.2403.03891 (2024).

# Supplemental Information

Filtering applied to compare TCGA with internal test sets:

In order to make appropriate comparisons among test sets, filters were applied to match test sets in terms of histology, tissue site, etc. TCGA samples consist of surgical resections of primary tumors. Histopathology algorithm performance can vary based on tissue site, so when comparing results on temporal and externally stained test sets to those on TCGA to inform generalization, results were compared using the subgroup of samples that were resected from the primary sites accordingly. In these comparisons, the test sets were also restricted to the most frequently occurring common histologies to further allow for more direct comparisons: i.e. transitional cell carcinoma and papillary transitional cell carcinoma in bladder cancer, infiltrating duct carcinoma and invasive ductal carcinoma in breast cancer, adenocarcinoma in colorectal cancer, and endometrioid adenocarcinoma, serous cystadenocarcinoma and papillary serous cystadenocarcinoma in endometrial cancer.

# Figures

**Figure 1**: Overall schematic of study. Multi-task models were trained to predict DNA alterations from H&E images in cross-validation in colorectal, breast, bladder and endometrial cancer. These models were evaluated on temporal holdout, external slides, and TCGA test sets.

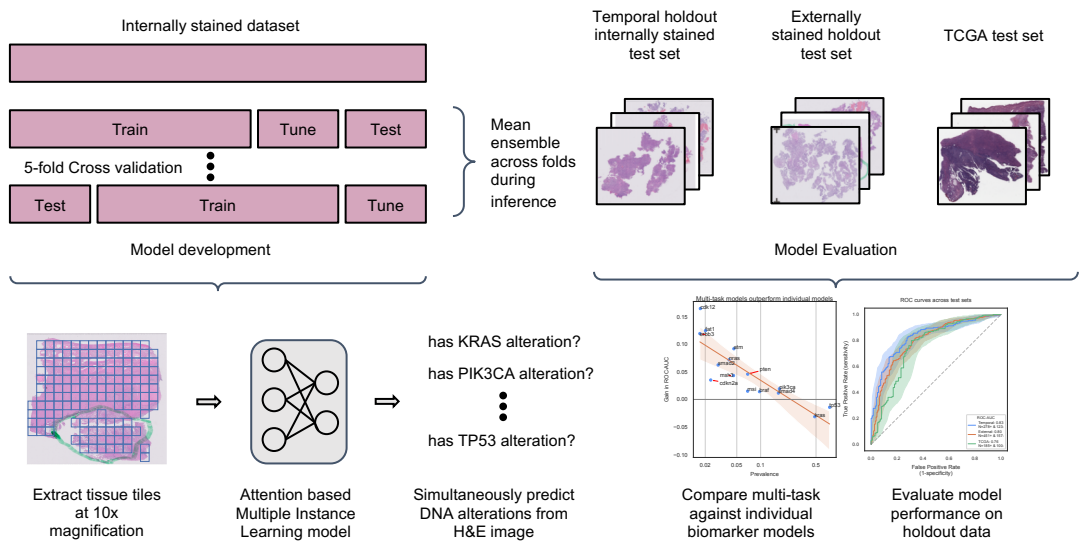

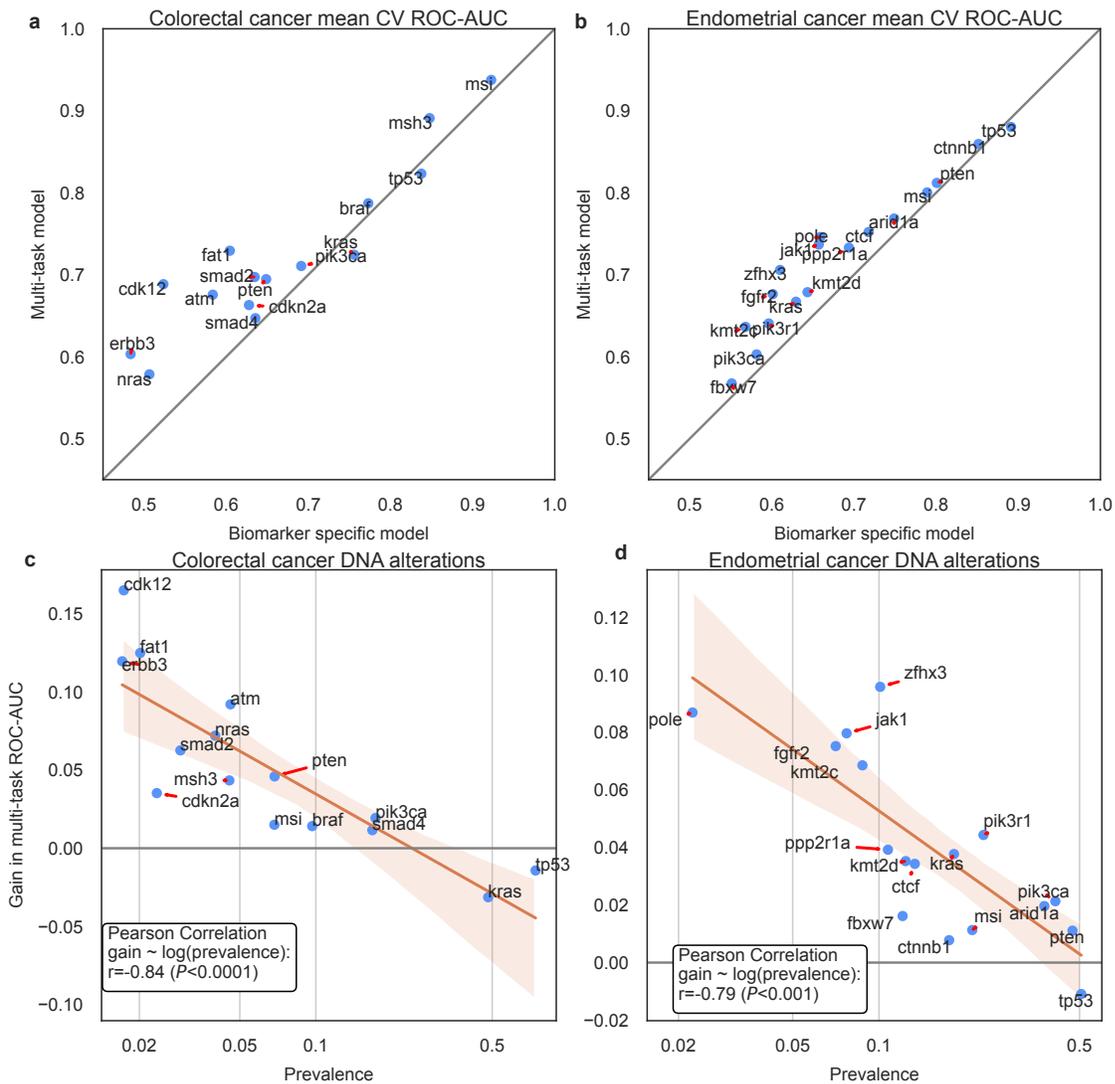

**Figure 2**: Comparison of multi-task with biomarker-specific models over cross-validation metrics. Scatterplots comparing multi-task and biomarker-specific model ROC-AUC in (a) colorectal cancer and (b) endometrial cancer. Scatterplot showing gain in multi-task model over single-task model ROC-AUC against prevalence (c) colorectal cancer and (d) endometrial cancer. Multi-task models achieved higher ROC-AUCs and the increase in ROC-AUC was correlated with prevalence.

**Figure 3**: Comparison of multi-task with biomarker-specific models over test set performance. Scatterplot comparing multi-task (blue markers) and single task (orange markers) model in terms of temporal holdout test set ROC-AUC on different biomarkers in (a) colorectal cancer and (b) endometrial cancer; externally stained slides test set in (c) colorectal cancer and (d) endometrial cancer; and TCGA test set in (e) colorectal cancer and (f) endometrial cancer. Multi-task models achieved higher holdout test set ROC-AUCs than biomarker-specific models on average.

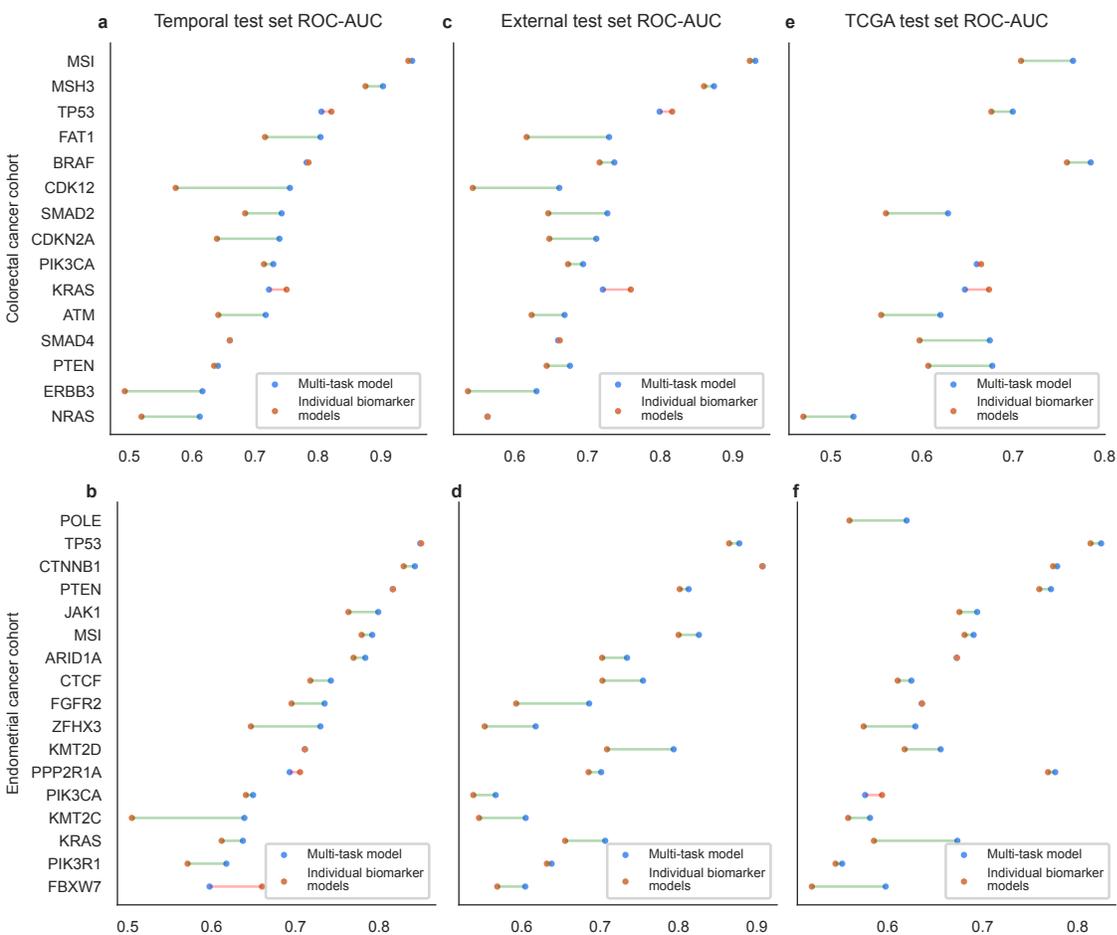

**Figure 4**: Relationship between target prevalence and multi-task model performance in four cancer types. Scatterplot shows mean cross-validation test folds ROC-AUC for different biomarkers against prevalence in (a) colorectal cancer, (b) endometrial cancer, (c) breast cancer, and (d) bladder cancer.

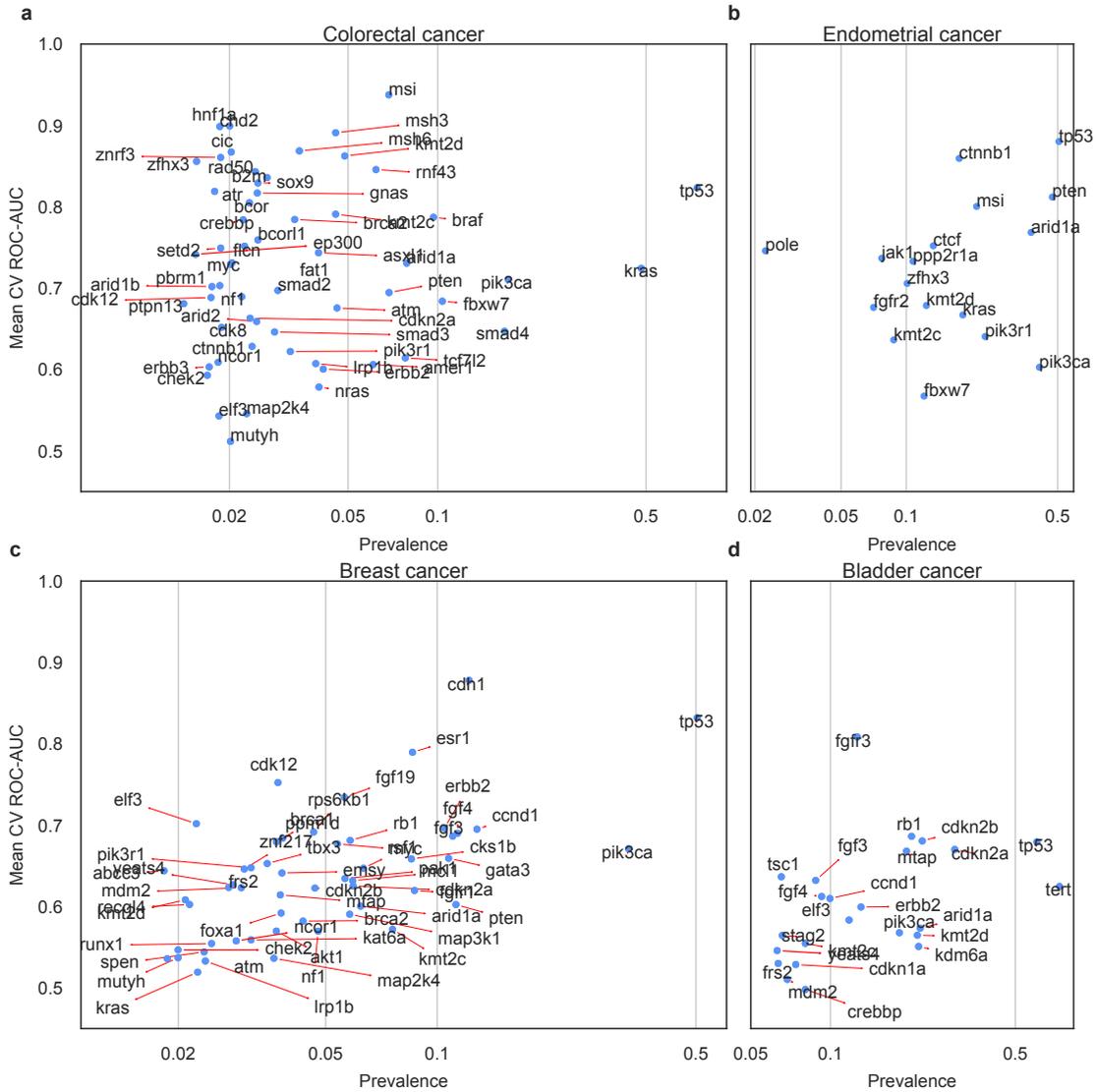

**Figure 5**: Multi-task models performance by target on test sets. Scatterplots showing multi-task model ROC-AUC on mean test folds (blue markers), temporal holdout test set (orange markers), externally stained test set (green markers) and TCGA test set (pink markers) in (a) colorectal cancer, (b) endometrial cancer, (c) breast cancer, and (d) bladder cancer.

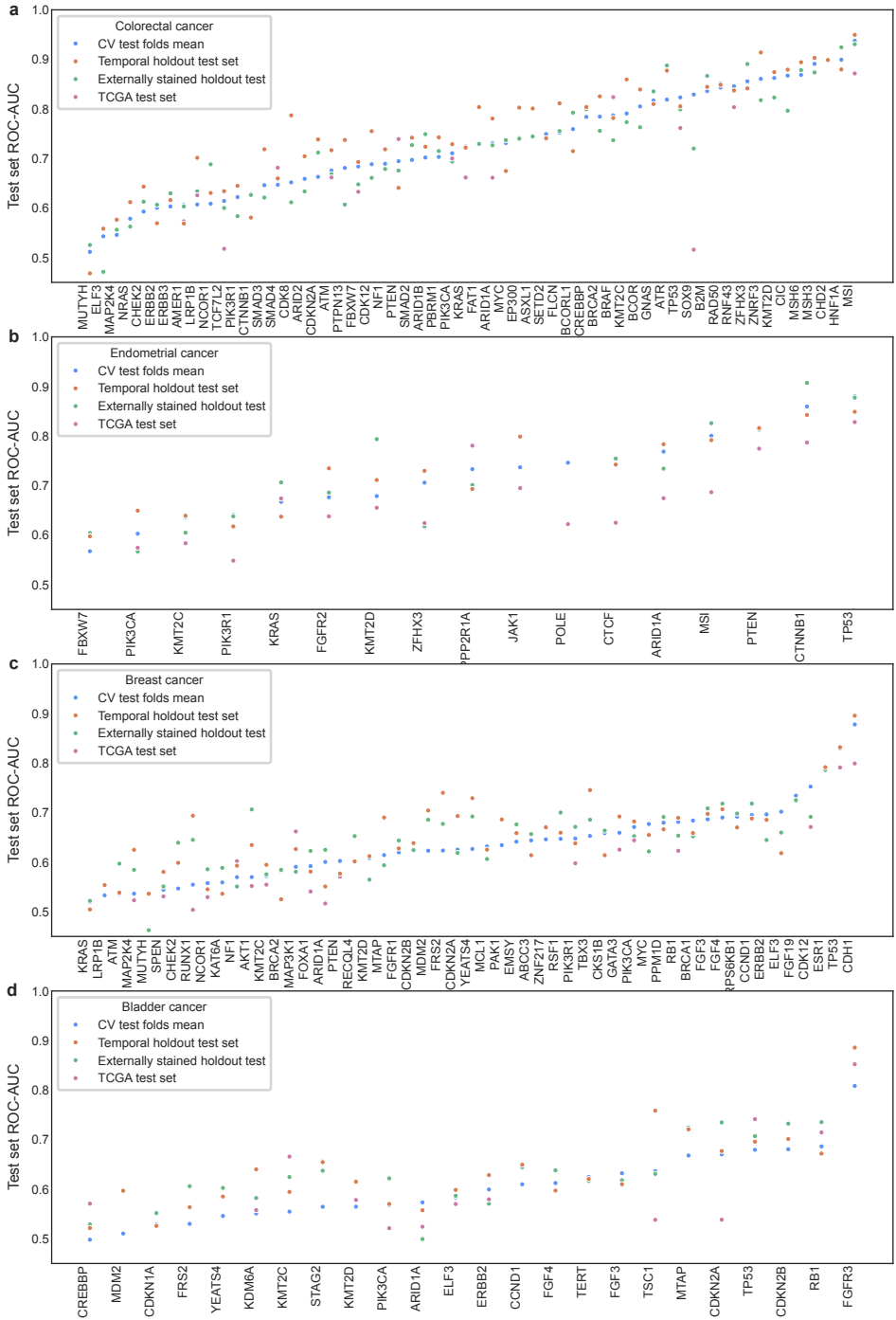

**Figure 6**: Comparing histology classes among high-attention tiles. Stripplot showing histology tile classes distribution among all tiles (blue markers) and among high-attention tiles (orange markers) defined as 10% of the tiles with highest attention scores in (a) colorectal cancer and (b) breast cancer. Tumor tiles were overrepresented among high-attention tiles than among all tiles.

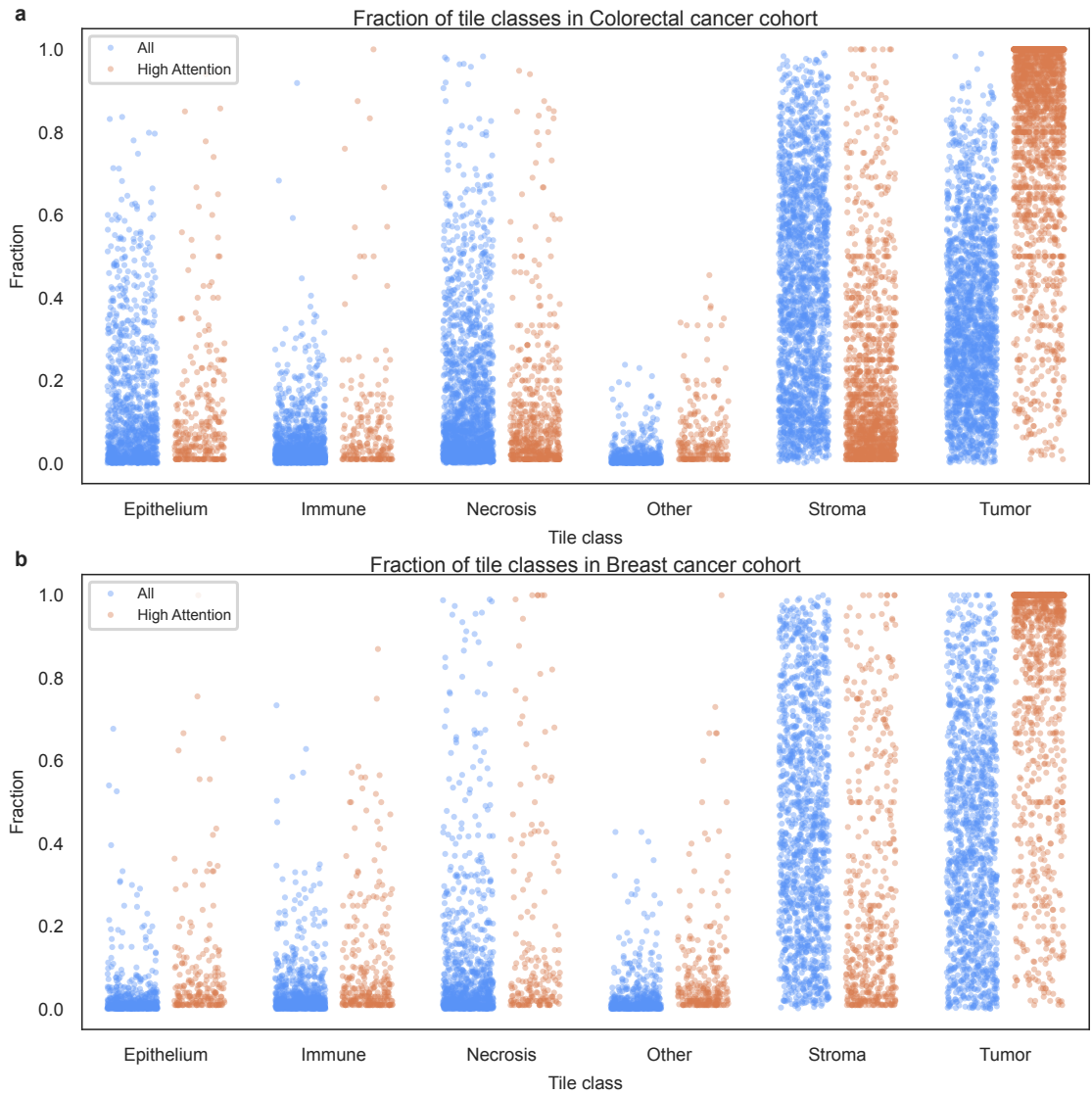

**Figure 7**: Comparing attention scores against tumor annotations. Receiver operating characteristic (ROC) curves were generated between attention scores and tumor annotations done by board-certified pathologists in (a) colorectal cancer and (b) breast cancer. Overlays of attention score heatmap and tumor annotation indicate a strong overlap in (c) colorectal cancer and (d) breast cancer.

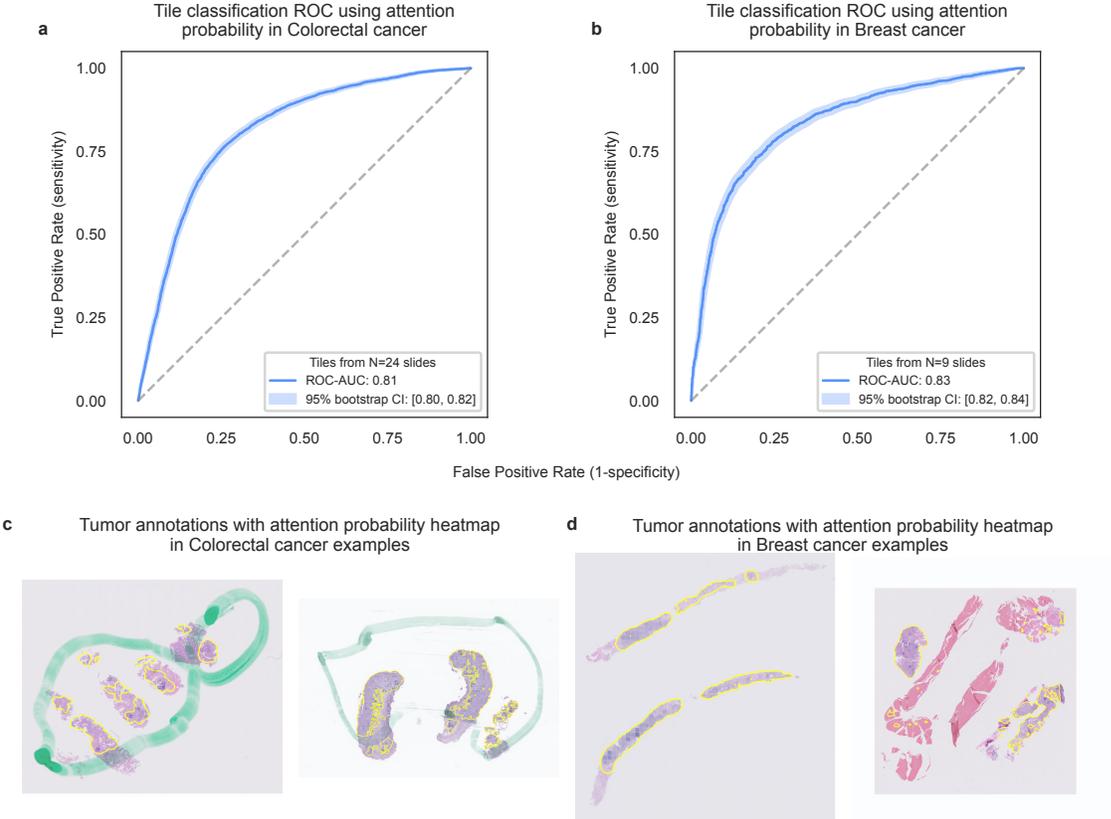

**Figure 8**: Multi-task model embeddings on downstream tasks. ROC curves indicate linear model performance on predicting cancer grade using multi-task model embeddings in (a) colorectal cancer, (b) endometrial cancer, (c) breast cancer, and (d) bladder cancer. Similar ROC curves for predicting primary tissue site using embeddings in (e) colorectal cancer, (g) endometrial cancer, (g) breast cancer, and (h) bladder cancer. Multi-task model embeddings were able to adapt to downstream tasks that were not a part of training.

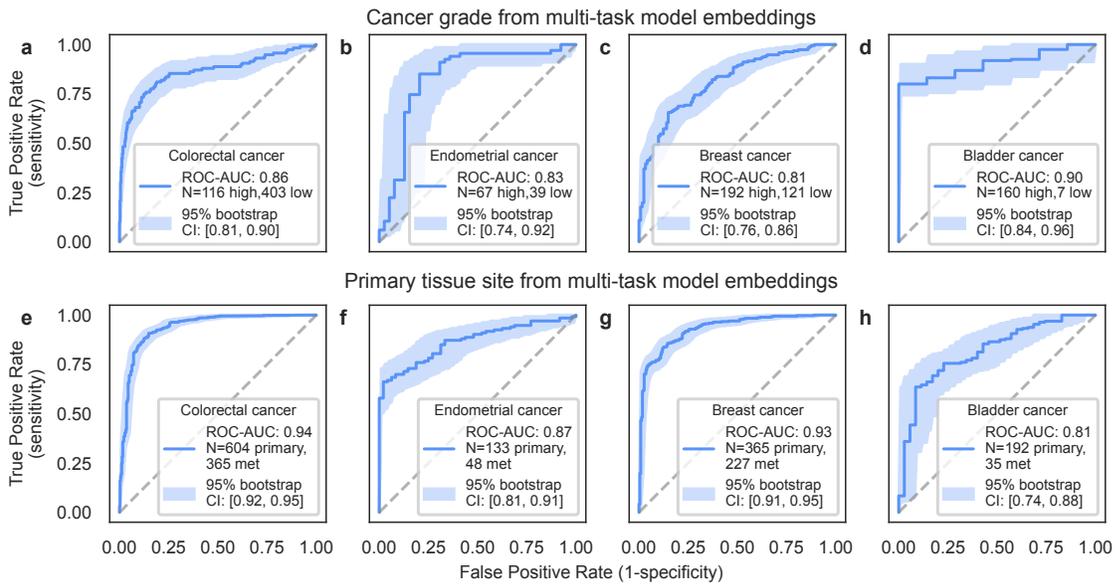

# Supplemental Figures

**Supplemental Figure 1:** Generalization to external and TCGA compared to in-house slides. Scatterplot comparing (a) performance on externally stained slides and (b) performance on TCGA slides against in-house temporal holdout slides. For this comparison, test sets were matched in terms of cohort characteristics as described in supplemental information.

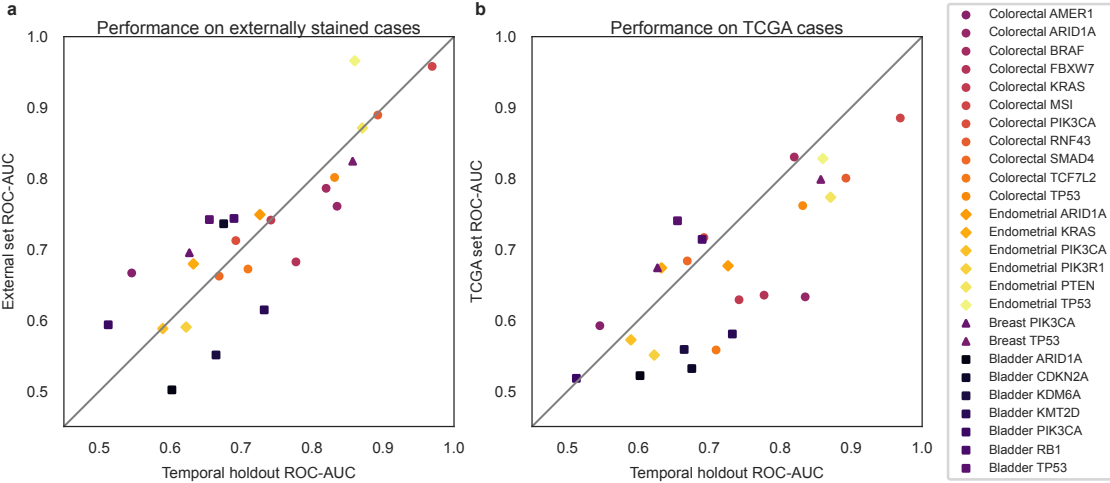

**Supplemental Figure 2:** ROC curves for a subset of targets over temporal holdout, externally stained, and TCGA test sets.

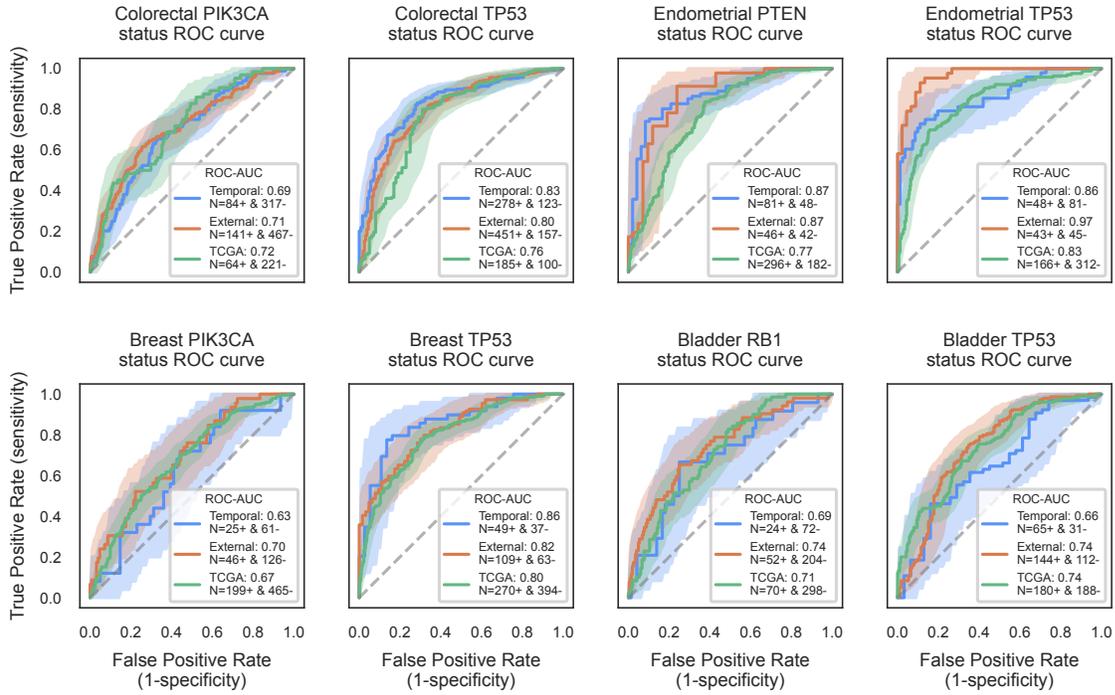

# Supplemental tables

**Supplemental Table 1:** Colorectal cancer cohort characteristics. Table showing patient characteristics in terms of demographics, stage, target labels, etc. in colorectal cancer cohort across model development, temporal holdout test set and externally stained test set.

| Characteristic | Model development set, N = 8,186[1] | Temporal holdout set, N = 2,037[1] | Externally stained holdout set, N = 2,660[1] |
|---|---|---|---|
| **Gender** | | | |
| Female | 3,530 (43%) | 877 (43%) | 1,148 (44%) |
| Male | 4,610 (57%) | 1,155 (57%) | 1,480 (56%) |
| Unknown | 46 | 5 | 32 |
| **Race** | | | |
| American Indian or Alaska Native | 34 (0.8%) | 5 (0.7%) | 7 (0.7%) |
| Asian | 167 (3.8%) | 26 (3.6%) | 60 (5.7%) |
| Black or African American | 604 (14%) | 68 (9.4%) | 148 (14%) |
| Native Hawaiian or Other Pacific Islander | 6 (0.1%) | 2 (0.3%) | 3 (0.3%) |
| Other Race | 247 (5.5%) | 41 (5.7%) | 71 (6.7%) |
| White | 3,395 (76%) | 579 (80%) | 768 (73%) |
| Unknown | 3,733 | 1,316 | 1,603 |
| **Stage** | | | |
| Stage 1 | 39 (0.6%) | 9 (0.6%) | 6 (0.3%) |
| Stage 2 | 210 (3.1%) | 55 (3.8%) | 47 (2.3%) |
| Stage 3 | 482 (7.2%) | 131 (9.0%) | 134 (6.5%) |
| Stage 4 | 5,976 (89%) | 1,259 (87%) | 1,882 (91%) |
| Unknown | 1,479 | 583 | 591 |
| **Tissue site** | | | |
| Colorectal | 4,654 (57%) | 1,048 (51%) | 1,392 (52%) |
| Lymph node | 172 (2.1%) | 43 (2.1%) | 54 (2.0%) |
| Other | 3,360 (41%) | 946 (46%) | 1,214 (46%) |
| **Specimen size** | | | |
| Large specimens | 3,835 (74%) | 695 (58%) | 1,360 (79%) |
| Small specimens | 1,369 (26%) | 503 (42%) | 361 (21%) |
| Unknown | 2,982 | 839 | 939 |
| **Scanner make** | | | |

| Characteristic | Model development set, N = 8,186[1] | Temporal holdout set, N = 2,037[1] | Externally stained holdout set, N = 2,660[1] |
|---|---|---|---|
| Aperio | 4,428 (54%) | 1,496 (73%) | 1,545 (58%) |
| Philips | 3,758 (46%) | 541 (27%) | 1,115 (42%) |
| **Has PTEN alteration** | | | |
| Negative | 7,624 (93%) | 1,870 (92%) | 2,478 (93%) |
| Positive | 562 (6.9%) | 167 (8.2%) | 182 (6.8%) |
| **Has PTPN13 alteration** | | | |
| Negative | 8,071 (99%) | 1,998 (98%) | 2,614 (98%) |
| Positive | 115 (1.4%) | 39 (1.9%) | 46 (1.7%) |
| **Has CDKN2A alteration** | | | |
| Negative | 7,994 (98%) | 1,994 (98%) | 2,601 (98%) |
| Positive | 192 (2.3%) | 43 (2.1%) | 59 (2.2%) |
| **Has SOX9 alteration** | | | |
| Negative | 7,982 (98%) | 2,037 (100%) | 2,595 (98%) |
| Positive | 204 (2.5%) | 0 (0%) | 65 (2.4%) |
| **Has FBXW7 alteration** | | | |
| Negative | 7,338 (90%) | 1,832 (90%) | 2,408 (91%) |
| Positive | 848 (10%) | 205 (10%) | 252 (9.5%) |
| **Has MSH3 alteration** | | | |
| Negative | 7,814 (95%) | 1,921 (94%) | 2,566 (96%) |
| Positive | 372 (4.5%) | 116 (5.7%) | 94 (3.5%) |
| **Has BRCA2 alteration** | | | |
| Negative | 7,915 (97%) | 1,952 (96%) | 2,578 (97%) |
| Positive | 271 (3.3%) | 85 (4.2%) | 82 (3.1%) |
| **Has CREBBP alteration** | | | |
| Negative | 8,004 (98%) | 1,967 (97%) | 2,603 (98%) |
| Positive | 182 (2.2%) | 70 (3.4%) | 57 (2.1%) |
| **Has MUTYH alteration** | | | |
| Negative | 8,021 (98%) | 1,988 (98%) | 2,604 (98%) |
| Positive | 165 (2.0%) | 49 (2.4%) | 56 (2.1%) |
| **Has CHD2 alteration** | | | |
| Negative | 8,034 (98%) | 2,002 (98%) | 2,615 (98%) |
| Positive | 152 (1.9%) | 35 (1.7%) | 45 (1.7%) |
| **Has BCOR alteration** | | | |
| Negative | 7,995 (98%) | 1,981 (97%) | 2,604 (98%) |
| Positive | 191 (2.3%) | 56 (2.7%) | 56 (2.1%) |
| **Has RAD50 alteration** | | | |
| Negative | 7,986 (98%) | 1,972 (97%) | 2,603 (98%) |
| Positive | 200 (2.4%) | 65 (3.2%) | 57 (2.1%) |

| Characteristic | Model development set, N = 8,186[1] | Temporal holdout set, N = 2,037[1] | Externally stained holdout set, N = 2,660[1] |
|---|---|---|---|
| **Has TP53 alteration** | | | |
| Negative | 2,108 (26%) | 554 (27%) | 686 (26%) |
| Positive | 6,078 (74%) | 1,483 (73%) | 1,974 (74%) |
| **Has PIK3CA alteration** | | | |
| Negative | 6,775 (83%) | 1,707 (84%) | 2,207 (83%) |
| Positive | 1,411 (17%) | 330 (16%) | 453 (17%) |
| **Has KMT2C alteration** | | | |
| Negative | 7,814 (95%) | 1,938 (95%) | 2,546 (96%) |
| Positive | 372 (4.5%) | 99 (4.9%) | 114 (4.3%) |
| **Has KMT2D alteration** | | | |
| Negative | 7,787 (95%) | 1,924 (94%) | 2,534 (95%) |
| Positive | 399 (4.9%) | 113 (5.5%) | 126 (4.7%) |
| **Has NCOR1 alteration** | | | |
| Negative | 8,036 (98%) | 1,996 (98%) | 2,618 (98%) |
| Positive | 150 (1.8%) | 41 (2.0%) | 42 (1.6%) |
| **Has ZNRF3 alteration** | | | |
| Negative | 8,033 (98%) | 1,986 (97%) | 2,615 (98%) |
| Positive | 153 (1.9%) | 51 (2.5%) | 45 (1.7%) |
| **Has KRAS alteration** | | | |
| Negative | 4,241 (52%) | 1,049 (51%) | 1,363 (51%) |
| Positive | 3,945 (48%) | 988 (49%) | 1,297 (49%) |
| **Has AMER1 alteration** | | | |
| Negative | 7,689 (94%) | 1,896 (93%) | 2,497 (94%) |
| Positive | 497 (6.1%) | 141 (6.9%) | 163 (6.1%) |
| **Has CDK12 alteration** | | | |
| Negative | 8,044 (98%) | 2,016 (99%) | 2,619 (98%) |
| Positive | 142 (1.7%) | 21 (1.0%) | 41 (1.5%) |
| **Has ERBB2 alteration** | | | |
| Negative | 7,848 (96%) | 1,941 (95%) | 2,533 (95%) |
| Positive | 338 (4.1%) | 96 (4.7%) | 127 (4.8%) |
| **Has HNF1A alteration** | | | |
| Negative | 8,022 (98%) | 1,994 (98%) | 2,624 (99%) |
| Positive | 164 (2.0%) | 43 (2.1%) | 36 (1.4%) |
| **Has SMAD4 alteration** | | | |
| Negative | 6,815 (83%) | 1,694 (83%) | 2,219 (83%) |
| Positive | 1,371 (17%) | 343 (17%) | 441 (17%) |
| **Has TCF7L2 alteration** | | | |
| Negative | 7,549 (92%) | 1,849 (91%) | 2,466 (93%) |

| Characteristic | Model development set, N = 8,186[1] | Temporal holdout set, N = 2,037[1] | Externally stained holdout set, N = 2,660[1] |
|---|---|---|---|
| Positive | 637 (7.8%) | 188 (9.2%) | 194 (7.3%) |
| **Has NRAS alteration** | | | |
| Negative | 7,859 (96%) | 1,943 (95%) | 2,551 (96%) |
| Positive | 327 (4.0%) | 94 (4.6%) | 109 (4.1%) |
| **Has ERBB3 alteration** | | | |
| Negative | 8,046 (98%) | 1,999 (98%) | 2,622 (99%) |
| Positive | 140 (1.7%) | 38 (1.9%) | 38 (1.4%) |
| **Has FAT1 alteration** | | | |
| Negative | 8,021 (98%) | 1,982 (97%) | 2,611 (98%) |
| Positive | 165 (2.0%) | 55 (2.7%) | 49 (1.8%) |
| **Has ASXL1 alteration** | | | |
| Negative | 7,860 (96%) | 1,929 (95%) | 2,563 (96%) |
| Positive | 326 (4.0%) | 108 (5.3%) | 97 (3.6%) |
| **Has PIK3R1 alteration** | | | |
| Negative | 7,924 (97%) | 1,958 (96%) | 2,559 (96%) |
| Positive | 262 (3.2%) | 79 (3.9%) | 101 (3.8%) |
| **Has MAP2K4 alteration** | | | |
| Negative | 7,999 (98%) | 1,991 (98%) | 2,604 (98%) |
| Positive | 187 (2.3%) | 46 (2.3%) | 56 (2.1%) |
| **Has MSI alteration** | | | |
| Negative | 7,625 (93%) | 1,864 (92%) | 2,513 (94%) |
| Positive | 561 (6.9%) | 173 (8.5%) | 147 (5.5%) |
| **Has B2M alteration** | | | |
| Negative | 7,967 (97%) | 1,983 (97%) | 2,602 (98%) |
| Positive | 219 (2.7%) | 54 (2.7%) | 58 (2.2%) |
| **Has ATR alteration** | | | |
| Negative | 8,040 (98%) | 1,987 (98%) | 2,619 (98%) |
| Positive | 146 (1.8%) | 50 (2.5%) | 41 (1.5%) |
| **Has EP300 alteration** | | | |
| Negative | 8,060 (98%) | 1,990 (98%) | 2,627 (99%) |
| Positive | 126 (1.5%) | 47 (2.3%) | 33 (1.2%) |
| **Has PBRM1 alteration** | | | |
| Negative | 8,034 (98%) | 1,998 (98%) | 2,617 (98%) |
| Positive | 152 (1.9%) | 39 (1.9%) | 43 (1.6%) |
| **Has MYC alteration** | | | |
| Negative | 8,019 (98%) | 2,008 (99%) | 2,626 (99%) |
| Positive | 167 (2.0%) | 29 (1.4%) | 34 (1.3%) |
| **Has ARID1A alteration** | | | |

| Characteristic | Model development set, N = 8,186[1] | Temporal holdout set, N = 2,037[1] | Externally stained holdout set, N = 2,660[1] |
|---|---|---|---|
| Negative | 7,542 (92%) | 1,856 (91%) | 2,475 (93%) |
| Positive | 644 (7.9%) | 181 (8.9%) | 185 (7.0%) |
| **Has MSH6 alteration** | | | |
| Negative | 7,905 (97%) | 1,962 (96%) | 2,581 (97%) |
| Positive | 281 (3.4%) | 75 (3.7%) | 79 (3.0%) |
| **Has BCORL1 alteration** | | | |
| Negative | 7,982 (98%) | 1,971 (97%) | 2,610 (98%) |
| Positive | 204 (2.5%) | 66 (3.2%) | 50 (1.9%) |
| **Has SETD2 alteration** | | | |
| Negative | 8,033 (98%) | 2,002 (98%) | 2,627 (99%) |
| Positive | 153 (1.9%) | 35 (1.7%) | 33 (1.2%) |
| **Has CDK8 alteration** | | | |
| Negative | 8,032 (98%) | 2,015 (99%) | 2,614 (98%) |
| Positive | 154 (1.9%) | 22 (1.1%) | 46 (1.7%) |
| **Has ATM alteration** | | | |
| Negative | 7,810 (95%) | 1,913 (94%) | 2,525 (95%) |
| Positive | 376 (4.6%) | 124 (6.1%) | 135 (5.1%) |
| **Has BRAF alteration** | | | |
| Negative | 7,394 (90%) | 1,831 (90%) | 2,434 (92%) |
| Positive | 792 (9.7%) | 206 (10%) | 226 (8.5%) |
| **Has ARID2 alteration** | | | |
| Negative | 7,984 (98%) | 1,980 (97%) | 2,602 (98%) |
| Positive | 202 (2.5%) | 57 (2.8%) | 58 (2.2%) |
| **Has CIC alteration** | | | |
| Negative | 8,020 (98%) | 1,985 (97%) | 2,612 (98%) |
| Positive | 166 (2.0%) | 52 (2.6%) | 48 (1.8%) |
| **Has ELF3 alteration** | | | |
| Negative | 8,035 (98%) | 1,999 (98%) | 2,616 (98%) |
| Positive | 151 (1.8%) | 38 (1.9%) | 44 (1.7%) |
| **Has SMAD3 alteration** | | | |
| Negative | 7,954 (97%) | 1,981 (97%) | 2,575 (97%) |
| Positive | 232 (2.8%) | 56 (2.7%) | 85 (3.2%) |
| **Has CHEK2 alteration** | | | |
| Negative | 8,048 (98%) | 2,015 (99%) | 2,610 (98%) |
| Positive | 138 (1.7%) | 22 (1.1%) | 50 (1.9%) |
| **Has NF1 alteration** | | | |
| Negative | 8,006 (98%) | 1,983 (97%) | 2,608 (98%) |
| Positive | 180 (2.2%) | 54 (2.7%) | 52 (2.0%) |

| Characteristic | Model development set, N = 8,186[1] | Temporal holdout set, N = 2,037[1] | Externally stained holdout set, N = 2,660[1] |
|---|---|---|---|
| **Has LRP1B alteration** | | | |
| Negative | 7,867 (96%) | 1,945 (95%) | 2,551 (96%) |
| Positive | 319 (3.9%) | 92 (4.5%) | 109 (4.1%) |
| **Has RNF43 alteration** | | | |
| Negative | 7,678 (94%) | 1,882 (92%) | 2,523 (95%) |
| Positive | 508 (6.2%) | 155 (7.6%) | 137 (5.2%) |
| **Has ARID1B alteration** | | | |
| Negative | 8,043 (98%) | 1,990 (98%) | 2,615 (98%) |
| Positive | 143 (1.7%) | 47 (2.3%) | 45 (1.7%) |
| **Has FLCN alteration** | | | |
| Negative | 8,002 (98%) | 1,979 (97%) | 2,609 (98%) |
| Positive | 184 (2.2%) | 58 (2.8%) | 51 (1.9%) |
| **Has CTNNB1 alteration** | | | |
| Negative | 7,991 (98%) | 1,991 (98%) | 2,590 (97%) |
| Positive | 195 (2.4%) | 46 (2.3%) | 70 (2.6%) |
| **Has ZFHX3 alteration** | | | |
| Negative | 8,059 (98%) | 1,998 (98%) | 2,630 (99%) |
| Positive | 127 (1.6%) | 39 (1.9%) | 30 (1.1%) |
| **Has SMAD2 alteration** | | | |
| Negative | 7,948 (97%) | 1,959 (96%) | 2,584 (97%) |
| Positive | 238 (2.9%) | 78 (3.8%) | 76 (2.9%) |
| **Has GNAS alteration** | | | |
| Negative | 7,983 (98%) | 1,991 (98%) | 2,595 (98%) |
| Positive | 203 (2.5%) | 46 (2.3%) | 65 (2.4%) |

[1] n (%)

**Supplemental Table 2:** Breast cancer cohort characteristics. Table showing patient characteristics in terms of demographics, stage, target labels, etc. in breast cancer cohort across model development, temporal holdout test set and externally stained test set.

| Characteristic | Model development set, N = 5,411[1] | Temporal holdout set, N = 1,356[1] | Externally stained holdout set, N = 2,038[1] |
|---|---|---|---|
| **Gender** | | | |
|   Female | 5,322 (99%) | 1,337 (99%) | 1,972 (99%) |
|   Male | 62 (1.2%) | 19 (1.4%) | 20 (1.0%) |
|   Unknown | 27 | 0 | 46 |
| **Race** | | | |
|   American Indian or Alaska Native | 16 (0.5%) | 4 (0.7%) | 2 (0.2%) |
|   Asian | 129 (4.1%) | 19 (3.5%) | 57 (5.8%) |
|   Black or African American | 452 (14%) | 65 (12%) | 159 (16%) |
|   Native Hawaiian or Other Pacific Islander | 6 (0.2%) | 1 (0.2%) | 1 (0.1%) |
|   Other Race | 159 (5.1%) | 28 (5.1%) | 61 (6.2%) |
|   White | 2,367 (76%) | 428 (79%) | 709 (72%) |
|   Unknown | 2,282 | 811 | 1,049 |
| **Stage** | | | |
|   Stage 0 | 0 (0%) | 0 (0%) | 1 (<0.1%) |
|   Stage 1 | 92 (2.1%) | 25 (2.7%) | 30 (1.9%) |
|   Stage 2 | 168 (3.8%) | 52 (5.5%) | 82 (5.3%) |
|   Stage 3 | 222 (5.0%) | 51 (5.4%) | 78 (5.1%) |
|   Stage 4 | 3,951 (89%) | 814 (86%) | 1,350 (88%) |
|   Unknown | 978 | 414 | 497 |
| **Tissue site** | | | |
|   Breast | 2,419 (45%) | 504 (37%) | 811 (40%) |
|   Lymph node | 568 (10%) | 135 (10.0%) | 201 (9.9%) |
|   Other | 2,424 (45%) | 717 (53%) | 1,026 (50%) |
| **Specimen size** | | | |
|   Large specimens | 1,653 (39%) | 263 (26%) | 633 (40%) |
|   Small specimens | 2,542 (61%) | 732 (74%) | 932 (60%) |
|   Unknown | 1,216 | 361 | 473 |
| **Scanner make** | | | |
|   Aperio | 2,697 (50%) | 963 (71%) | 1,091 (54%) |
|   Philips | 2,714 (50%) | 393 (29%) | 947 (46%) |
| **Has RSF1 alteration** | | | |

| Characteristic | Model development set, N = 5,411[1] | Temporal holdout set, N = 1,356[1] | Externally stained holdout set, N = 2,038[1] |
|---|---|---|---|
| Negative | 5,069 (94%) | 1,269 (94%) | 1,900 (93%) |
| Positive | 342 (6.3%) | 87 (6.4%) | 138 (6.8%) |
| **Has MAP3K1 alteration** | | | |
| Negative | 5,097 (94%) | 1,267 (93%) | 1,938 (95%) |
| Positive | 314 (5.8%) | 89 (6.6%) | 100 (4.9%) |
| **Has ERBB2 alteration** | | | |
| Negative | 4,845 (90%) | 1,227 (90%) | 1,856 (91%) |
| Positive | 566 (10%) | 129 (9.5%) | 182 (8.9%) |
| **Has ARID1A alteration** | | | |
| Negative | 5,075 (94%) | 1,282 (95%) | 1,930 (95%) |
| Positive | 336 (6.2%) | 74 (5.5%) | 108 (5.3%) |
| **Has PPM1D alteration** | | | |
| Negative | 5,212 (96%) | 1,306 (96%) | 1,970 (97%) |
| Positive | 199 (3.7%) | 50 (3.7%) | 68 (3.3%) |
| **Has MTAP alteration** | | | |
| Negative | 5,207 (96%) | 1,326 (98%) | 1,968 (97%) |
| Positive | 204 (3.8%) | 30 (2.2%) | 70 (3.4%) |
| **Has ELF3 alteration** | | | |
| Negative | 5,290 (98%) | 1,334 (98%) | 2,003 (98%) |
| Positive | 121 (2.2%) | 22 (1.6%) | 35 (1.7%) |
| **Has CKS1B alteration** | | | |
| Negative | 4,950 (91%) | 1,278 (94%) | 1,926 (95%) |
| Positive | 461 (8.5%) | 78 (5.8%) | 112 (5.5%) |
| **Has YEATS4 alteration** | | | |
| Negative | 5,259 (97%) | 1,321 (97%) | 1,970 (97%) |
| Positive | 152 (2.8%) | 35 (2.6%) | 68 (3.3%) |
| **Has CDK12 alteration** | | | |
| Negative | 5,210 (96%) | 1,353 (100%) | 1,986 (97%) |
| Positive | 201 (3.7%) | 3 (0.2%) | 52 (2.6%) |
| **Has FRS2 alteration** | | | |
| Negative | 5,263 (97%) | 1,324 (98%) | 1,972 (97%) |
| Positive | 148 (2.7%) | 32 (2.4%) | 66 (3.2%) |
| **Has ZNF217 alteration** | | | |
| Negative | 5,248 (97%) | 1,319 (97%) | 1,966 (96%) |
| Positive | 163 (3.0%) | 37 (2.7%) | 72 (3.5%) |
| **Has PAK1 alteration** | | | |
| Negative | 5,106 (94%) | 1,279 (94%) | 1,926 (95%) |
| Positive | 305 (5.6%) | 77 (5.7%) | 112 (5.5%) |

| Characteristic | Model development set, N = 5,411[1] | Temporal holdout set, N = 1,356[1] | Externally stained holdout set, N = 2,038[1] |
|---|---|---|---|
| **Has FGFR1 alteration** | | | |
| Negative | 4,941 (91%) | 1,245 (92%) | 1,843 (90%) |
| Positive | 470 (8.7%) | 111 (8.2%) | 195 (9.6%) |
| **Has KMT2C alteration** | | | |
| Negative | 5,001 (92%) | 1,218 (90%) | 1,897 (93%) |
| Positive | 410 (7.6%) | 138 (10%) | 141 (6.9%) |
| **Has LRP1B alteration** | | | |
| Negative | 5,283 (98%) | 1,330 (98%) | 1,985 (97%) |
| Positive | 128 (2.4%) | 26 (1.9%) | 53 (2.6%) |
| **Has KAT6A alteration** | | | |
| Negative | 5,241 (97%) | 1,314 (97%) | 1,980 (97%) |
| Positive | 170 (3.1%) | 42 (3.1%) | 58 (2.8%) |
| **Has NF1 alteration** | | | |
| Negative | 5,153 (95%) | 1,304 (96%) | 1,936 (95%) |
| Positive | 258 (4.8%) | 52 (3.8%) | 102 (5.0%) |
| **Has RB1 alteration** | | | |
| Negative | 5,096 (94%) | 1,274 (94%) | 1,905 (93%) |
| Positive | 315 (5.8%) | 82 (6.0%) | 133 (6.5%) |
| **Has ATM alteration** | | | |
| Negative | 5,310 (98%) | 1,322 (97%) | 1,996 (98%) |
| Positive | 101 (1.9%) | 34 (2.5%) | 42 (2.1%) |
| **Has ABCC3 alteration** | | | |
| Negative | 5,312 (98%) | 1,331 (98%) | 1,998 (98%) |
| Positive | 99 (1.8%) | 25 (1.8%) | 40 (2.0%) |
| **Has FGF4 alteration** | | | |
| Negative | 4,797 (89%) | 1,190 (88%) | 1,820 (89%) |
| Positive | 614 (11%) | 166 (12%) | 218 (11%) |
| **Has KRAS alteration** | | | |
| Negative | 5,289 (98%) | 1,330 (98%) | 1,992 (98%) |
| Positive | 122 (2.3%) | 26 (1.9%) | 46 (2.3%) |
| **Has KMT2D alteration** | | | |
| Negative | 5,298 (98%) | 1,326 (98%) | 1,993 (98%) |
| Positive | 113 (2.1%) | 30 (2.2%) | 45 (2.2%) |
| **Has ESR1 alteration** | | | |
| Negative | 4,947 (91%) | 1,221 (90%) | 1,839 (90%) |
| Positive | 464 (8.6%) | 135 (10.0%) | 199 (9.8%) |
| **Has CDKN2B alteration** | | | |
| Negative | 5,158 (95%) | 1,298 (96%) | 1,952 (96%) |

| Characteristic | Model development set, N = 5,411[1] | Temporal holdout set, N = 1,356[1] | Externally stained holdout set, N = 2,038[1] |
|---|---|---|---|
| Positive | 253 (4.7%) | 58 (4.3%) | 86 (4.2%) |
| **Has TP53 alteration** | | | |
| Negative | 2,684 (50%) | 675 (50%) | 979 (48%) |
| Positive | 2,727 (50%) | 681 (50%) | 1,059 (52%) |
| **Has BRCA1 alteration** | | | |
| Negative | 5,204 (96%) | 1,319 (97%) | 1,968 (97%) |
| Positive | 207 (3.8%) | 37 (2.7%) | 70 (3.4%) |
| **Has PIK3R1 alteration** | | | |
| Negative | 5,241 (97%) | 1,319 (97%) | 1,975 (97%) |
| Positive | 170 (3.1%) | 37 (2.7%) | 63 (3.1%) |
| **Has MCL1 alteration** | | | |
| Negative | 5,091 (94%) | 1,320 (97%) | 1,936 (95%) |
| Positive | 320 (5.9%) | 36 (2.7%) | 102 (5.0%) |
| **Has RECQL4 alteration** | | | |
| Negative | 5,295 (98%) | 1,324 (98%) | 1,991 (98%) |
| Positive | 116 (2.1%) | 32 (2.4%) | 47 (2.3%) |
| **Has NCOR1 alteration** | | | |
| Negative | 5,256 (97%) | 1,309 (97%) | 1,980 (97%) |
| Positive | 155 (2.9%) | 47 (3.5%) | 58 (2.8%) |
| **Has RPS6KB1 alteration** | | | |
| Negative | 5,160 (95%) | 1,305 (96%) | 1,955 (96%) |
| Positive | 251 (4.6%) | 51 (3.8%) | 83 (4.1%) |
| **Has SPEN alteration** | | | |
| Negative | 5,284 (98%) | 1,322 (97%) | 2,003 (98%) |
| Positive | 127 (2.3%) | 34 (2.5%) | 35 (1.7%) |
| **Has FGF19 alteration** | | | |
| Negative | 5,107 (94%) | 1,356 (100%) | 1,949 (96%) |
| Positive | 304 (5.6%) | 0 (0%) | 89 (4.4%) |
| **Has MYC alteration** | | | |
| Negative | 5,120 (95%) | 1,291 (95%) | 1,949 (96%) |
| Positive | 291 (5.4%) | 65 (4.8%) | 89 (4.4%) |
| **Has RUNX1 alteration** | | | |
| Negative | 5,278 (98%) | 1,330 (98%) | 2,000 (98%) |
| Positive | 133 (2.5%) | 26 (1.9%) | 38 (1.9%) |
| **Has AKT1 alteration** | | | |
| Negative | 5,212 (96%) | 1,304 (96%) | 1,955 (96%) |
| Positive | 199 (3.7%) | 52 (3.8%) | 83 (4.1%) |
| **Has CDKN2A alteration** | | | |

| Characteristic | Model development set, N = 5,411[1] | Temporal holdout set, N = 1,356[1] | Externally stained holdout set, N = 2,038[1] |
|---|---|---|---|
| Negative | 5,090 (94%) | 1,290 (95%) | 1,939 (95%) |
| Positive | 321 (5.9%) | 66 (4.9%) | 99 (4.9%) |
| **Has PIK3CA alteration** | | | |
| Negative | 3,632 (67%) | 894 (66%) | 1,392 (68%) |
| Positive | 1,779 (33%) | 462 (34%) | 646 (32%) |
| **Has CCND1 alteration** | | | |
| Negative | 4,718 (87%) | 1,180 (87%) | 1,787 (88%) |
| Positive | 693 (13%) | 176 (13%) | 251 (12%) |
| **Has MUTYH alteration** | | | |
| Negative | 5,303 (98%) | 1,331 (98%) | 1,998 (98%) |
| Positive | 108 (2.0%) | 25 (1.8%) | 40 (2.0%) |
| **Has EMSY alteration** | | | |
| Negative | 5,205 (96%) | 1,293 (95%) | 1,955 (96%) |
| Positive | 206 (3.8%) | 63 (4.6%) | 83 (4.1%) |
| **Has GATA3 alteration** | | | |
| Negative | 4,830 (89%) | 1,211 (89%) | 1,806 (89%) |
| Positive | 581 (11%) | 145 (11%) | 232 (11%) |
| **Has MDM2 alteration** | | | |
| Negative | 5,251 (97%) | 1,319 (97%) | 1,973 (97%) |
| Positive | 160 (3.0%) | 37 (2.7%) | 65 (3.2%) |
| **Has FOXA1 alteration** | | | |
| Negative | 5,206 (96%) | 1,302 (96%) | 1,955 (96%) |
| Positive | 205 (3.8%) | 54 (4.0%) | 83 (4.1%) |
| **Has CDH1 alteration** | | | |
| Negative | 4,751 (88%) | 1,183 (87%) | 1,803 (88%) |
| Positive | 660 (12%) | 173 (13%) | 235 (12%) |
| **Has BRCA2 alteration** | | | |
| Negative | 5,176 (96%) | 1,302 (96%) | 1,953 (96%) |
| Positive | 235 (4.3%) | 54 (4.0%) | 85 (4.2%) |
| **Has TBX3 alteration** | | | |
| Negative | 5,223 (97%) | 1,296 (96%) | 1,986 (97%) |
| Positive | 188 (3.5%) | 60 (4.4%) | 52 (2.6%) |
| **Has MAP2K4 alteration** | | | |
| Negative | 5,215 (96%) | 1,296 (96%) | 1,966 (96%) |
| Positive | 196 (3.6%) | 60 (4.4%) | 72 (3.5%) |
| **Has FGF3 alteration** | | | |
| Negative | 4,815 (89%) | 1,188 (88%) | 1,823 (89%) |
| Positive | 596 (11%) | 168 (12%) | 215 (11%) |

| Characteristic | Model development set, N = 5,411[1] | Temporal holdout set, N = 1,356[1] | Externally stained holdout set, N = 2,038[1] |
|---|---|---|---|
| **Has CHEK2 alteration** | | | |
| Negative | 5,303 (98%) | 1,315 (97%) | 2,002 (98%) |
| Positive | 108 (2.0%) | 41 (3.0%) | 36 (1.8%) |
| **Has PTEN alteration** | | | |
| Negative | 4,803 (89%) | 1,227 (90%) | 1,798 (88%) |
| Positive | 608 (11%) | 129 (9.5%) | 240 (12%) |

[1] n (%)

**Supplemental Table 3:** Endometrial cancer cohort characteristics. Table showing patient characteristics in terms of demographics, stage, target labels, etc. in endometrial cancer cohort across model development, temporal holdout test set and externally stained test set.

| Characteristic | Model development set, N = 2,011[1] | Temporal holdout set, N = 500[1] | Externally stained holdout set, N = 389[1] |
|---|---|---|---|
| **Gender** | | | |
| Female | 1,982 (100%) | 498 (100%) | 378 (100%) |
| Male | 3 (0.2%) | 1 (0.2%) | 1 (0.3%) |
| Unknown | 26 | 1 | 10 |
| **Race** | | | |
| American Indian or Alaska Native | 5 (0.4%) | 1 (0.6%) | 2 (1.3%) |
| Asian | 34 (2.9%) | 8 (4.6%) | 8 (5.2%) |
| Black or African American | 167 (14%) | 16 (9.2%) | 32 (21%) |
| Native Hawaiian or Other Pacific Islander | 1 (<0.1%) | 1 (0.6%) | 0 (0%) |
| Other Race | 60 (5.2%) | 16 (9.2%) | 8 (5.2%) |
| White | 894 (77%) | 131 (76%) | 105 (68%) |
| Unknown | 850 | 327 | 234 |
| **Stage** | | | |
| Stage 0 | 1 (<0.1%) | 0 (0%) | 0 (0%) |
| Stage 1 | 158 (11%) | 25 (9.7%) | 10 (4.2%) |
| Stage 2 | 30 (2.2%) | 11 (4.3%) | 2 (0.8%) |
| Stage 3 | 102 (7.3%) | 25 (9.7%) | 8 (3.3%) |
| Stage 4 | 1,097 (79%) | 197 (76%) | 220 (92%) |
| Unknown | 623 | 242 | 149 |
| **Tissue site** | | | |
| Endometrium | 484 (24%) | 125 (25%) | 89 (23%) |
| Lymph node | 88 (4.4%) | 25 (5.0%) | 21 (5.4%) |
| Other | 1,439 (72%) | 350 (70%) | 279 (72%) |
| **Specimen size** | | | |
| Large specimens | 1,293 (84%) | 296 (84%) | 219 (80%) |
| Small specimens | 241 (16%) | 58 (16%) | 54 (20%) |
| Unknown | 477 | 146 | 116 |
| **Scanner make** | | | |
| Aperio | 1,016 (51%) | 358 (72%) | 205 (53%) |
| Philips | 995 (49%) | 142 (28%) | 184 (47%) |
| **Has KMT2D alteration** | | | |

| Characteristic | Model development set, N = 2,011[1] | Temporal holdout set, N = 500[1] | Externally stained holdout set, N = 389[1] |
|---|---|---|---|
| Negative | 1,762 (88%) | 428 (86%) | 345 (89%) |
| Positive | 249 (12%) | 72 (14%) | 44 (11%) |
| **Has JAK1 alteration** | | | |
| Negative | 1,856 (92%) | 461 (92%) | 373 (96%) |
| Positive | 155 (7.7%) | 39 (7.8%) | 16 (4.1%) |
| **Has PIK3R1 alteration** | | | |
| Negative | 1,546 (77%) | 393 (79%) | 304 (78%) |
| Positive | 465 (23%) | 107 (21%) | 85 (22%) |
| **Has KMT2C alteration** | | | |
| Negative | 1,835 (91%) | 444 (89%) | 350 (90%) |
| Positive | 176 (8.8%) | 56 (11%) | 39 (10%) |
| **Has PTEN alteration** | | | |
| Negative | 1,061 (53%) | 243 (49%) | 224 (58%) |
| Positive | 950 (47%) | 257 (51%) | 165 (42%) |
| **Has ZFHX3 alteration** | | | |
| Negative | 1,808 (90%) | 446 (89%) | 361 (93%) |
| Positive | 203 (10%) | 54 (11%) | 28 (7.2%) |
| **Has CTNNB1 alteration** | | | |
| Negative | 1,658 (82%) | 414 (83%) | 317 (81%) |
| Positive | 353 (18%) | 86 (17%) | 72 (19%) |
| **Has POLE alteration** | | | |
| Negative | 1,966 (98%) | 490 (98%) | 379 (97%) |
| Positive | 45 (2.2%) | 10 (2.0%) | 10 (2.6%) |
| **Has FGFR2 alteration** | | | |
| Negative | 1,869 (93%) | 455 (91%) | 361 (93%) |
| Positive | 142 (7.1%) | 45 (9.0%) | 28 (7.2%) |
| **Has CTCF alteration** | | | |
| Negative | 1,743 (87%) | 420 (84%) | 345 (89%) |
| Positive | 268 (13%) | 80 (16%) | 44 (11%) |
| **Has KRAS alteration** | | | |
| Negative | 1,644 (82%) | 411 (82%) | 319 (82%) |
| Positive | 367 (18%) | 89 (18%) | 70 (18%) |
| **Has PPP2R1A alteration** | | | |
| Negative | 1,795 (89%) | 446 (89%) | 338 (87%) |
| Positive | 216 (11%) | 54 (11%) | 51 (13%) |
| **Has ARID1A alteration** | | | |
| Negative | 1,253 (62%) | 300 (60%) | 244 (63%) |
| Positive | 758 (38%) | 200 (40%) | 145 (37%) |

| Characteristic | Model development set, N = 2,011[1] | Temporal holdout set, N = 500[1] | Externally stained holdout set, N = 389[1] |
|---|---|---|---|
| **Has MSI alteration** | | | |
|   Negative | 1,586 (79%) | 390 (78%) | 328 (84%) |
|   Positive | 425 (21%) | 110 (22%) | 61 (16%) |
| **Has PIK3CA alteration** | | | |
|   Negative | 1,183 (59%) | 306 (61%) | 240 (62%) |
|   Positive | 828 (41%) | 194 (39%) | 149 (38%) |
| **Has FBXW7 alteration** | | | |
|   Negative | 1,768 (88%) | 444 (89%) | 346 (89%) |
|   Positive | 243 (12%) | 56 (11%) | 43 (11%) |
| **Has TP53 alteration** | | | |
|   Negative | 992 (49%) | 263 (53%) | 187 (48%) |
|   Positive | 1,019 (51%) | 237 (47%) | 202 (52%) |

[1] n (%)

**Supplemental Table 4:** Bladder cancer cohort characteristics. Table showing patient characteristics in terms of demographics, stage, target labels, etc. in bladder cancer cohort across model development, temporal holdout test set and externally stained test set.

| Characteristic | Model development set, N = 1,717[1] | Temporal holdout set, N = 425[1] | Externally stained holdout set, N = 731[1] |
|---|---|---|---|
| **Gender** | | | |
|   Female | 493 (29%) | 123 (29%) | 187 (27%) |
|   Male | 1,222 (71%) | 302 (71%) | 516 (73%) |
|   Unknown | 2 | 0 | 28 |
| **Race** | | | |
|   American Indian or Alaska Native | 5 (0.5%) | 0 (0%) | 0 (0%) |
|   Asian | 25 (2.4%) | 4 (1.8%) | 27 (7.1%) |
|   Black or African American | 85 (8.0%) | 20 (8.8%) | 33 (8.7%) |
|   Other Race | 55 (5.2%) | 13 (5.7%) | 15 (3.9%) |
|   White | 887 (84%) | 191 (84%) | 306 (80%) |
|   Unknown | 660 | 197 | 350 |
| **Stage** | | | |
|   Stage 0 | 2 (0.2%) | 0 (0%) | 1 (0.3%) |
|   Stage 1 | 18 (1.5%) | 4 (1.4%) | 7 (1.8%) |
|   Stage 2 | 116 (9.5%) | 18 (6.4%) | 22 (5.8%) |
|   Stage 3 | 102 (8.4%) | 30 (11%) | 45 (12%) |
|   Stage 4 | 981 (80%) | 228 (81%) | 304 (80%) |
|   Unknown | 498 | 145 | 352 |
| **Tissue site** | | | |
|   Bladder | 1,015 (59%) | 229 (54%) | 447 (61%) |
|   Lymph node | 149 (8.7%) | 37 (8.7%) | 47 (6.4%) |
|   Other | 553 (32%) | 159 (37%) | 237 (32%) |
| **Specimen size** | | | |
|   Large specimens | 1,113 (82%) | 234 (72%) | 565 (91%) |
|   Small specimens | 242 (18%) | 91 (28%) | 56 (9.0%) |
|   Unknown | 362 | 100 | 110 |
| **Scanner make** | | | |
|   Aperio | 879 (51%) | 330 (78%) | 377 (52%) |
|   Philips | 838 (49%) | 95 (22%) | 354 (48%) |
| **Has TP53 alteration** | | | |
|   Negative | 681 (40%) | 162 (38%) | 320 (44%) |
|   Positive | 1,036 (60%) | 263 (62%) | 411 (56%) |

| Characteristic | Model development set, N = 1,717[1] | Temporal holdout set, N = 425[1] | Externally stained holdout set, N = 731[1] |
|---|---|---|---|
| **Has KDM6A alteration** | | | |
| Negative | 1,347 (78%) | 333 (78%) | 556 (76%) |
| Positive | 370 (22%) | 92 (22%) | 175 (24%) |
| **Has FGF4 alteration** | | | |
| Negative | 1,558 (91%) | 380 (89%) | 674 (92%) |
| Positive | 159 (9.3%) | 45 (11%) | 57 (7.8%) |
| **Has STAG2 alteration** | | | |
| Negative | 1,604 (93%) | 391 (92%) | 675 (92%) |
| Positive | 113 (6.6%) | 34 (8.0%) | 56 (7.7%) |
| **Has YEATS4 alteration** | | | |
| Negative | 1,609 (94%) | 388 (91%) | 666 (91%) |
| Positive | 108 (6.3%) | 37 (8.7%) | 65 (8.9%) |
| **Has CCND1 alteration** | | | |
| Negative | 1,546 (90%) | 387 (91%) | 665 (91%) |
| Positive | 171 (10.0%) | 38 (8.9%) | 66 (9.0%) |
| **Has CDKN2B alteration** | | | |
| Negative | 1,335 (78%) | 341 (80%) | 565 (77%) |
| Positive | 382 (22%) | 84 (20%) | 166 (23%) |
| **Has ARID1A alteration** | | | |
| Negative | 1,342 (78%) | 318 (75%) | 574 (79%) |
| Positive | 375 (22%) | 107 (25%) | 157 (21%) |
| **Has TSC1 alteration** | | | |
| Negative | 1,605 (93%) | 396 (93%) | 674 (92%) |
| Positive | 112 (6.5%) | 29 (6.8%) | 57 (7.8%) |
| **Has CDKN1A alteration** | | | |
| Negative | 1,590 (93%) | 392 (92%) | 678 (93%) |
| Positive | 127 (7.4%) | 33 (7.8%) | 53 (7.3%) |
| **Has KMT2D alteration** | | | |
| Negative | 1,351 (79%) | 322 (76%) | 572 (78%) |
| Positive | 366 (21%) | 103 (24%) | 159 (22%) |
| **Has FRS2 alteration** | | | |
| Negative | 1,608 (94%) | 388 (91%) | 666 (91%) |
| Positive | 109 (6.3%) | 37 (8.7%) | 65 (8.9%) |
| **Has CREBBP alteration** | | | |
| Negative | 1,579 (92%) | 390 (92%) | 666 (91%) |
| Positive | 138 (8.0%) | 35 (8.2%) | 65 (8.9%) |
| **Has MTAP alteration** | | | |
| Negative | 1,384 (81%) | 353 (83%) | 580 (79%) |

| Characteristic | Model development set, N = 1,717[1] | Temporal holdout set, N = 425[1] | Externally stained holdout set, N = 731[1] |
|---|---|---|---|
| Positive | 333 (19%) | 72 (17%) | 151 (21%) |
| **Has KMT2C alteration** | | | |
| Negative | 1,579 (92%) | 385 (91%) | 659 (90%) |
| Positive | 138 (8.0%) | 40 (9.4%) | 72 (9.8%) |
| **Has CDKN2A alteration** | | | |
| Negative | 1,210 (70%) | 317 (75%) | 508 (69%) |
| Positive | 507 (30%) | 108 (25%) | 223 (31%) |
| **Has ELF3 alteration** | | | |
| Negative | 1,515 (88%) | 382 (90%) | 643 (88%) |
| Positive | 202 (12%) | 43 (10%) | 88 (12%) |
| **Has FGF3 alteration** | | | |
| Negative | 1,566 (91%) | 386 (91%) | 679 (93%) |
| Positive | 151 (8.8%) | 39 (9.2%) | 52 (7.1%) |
| **Has FGFR3 alteration** | | | |
| Negative | 1,500 (87%) | 377 (89%) | 645 (88%) |
| Positive | 217 (13%) | 48 (11%) | 86 (12%) |
| **Has TERT alteration** | | | |
| Negative | 456 (27%) | 111 (26%) | 208 (28%) |
| Positive | 1,261 (73%) | 314 (74%) | 523 (72%) |
| **Has PIK3CA alteration** | | | |
| Negative | 1,404 (82%) | 352 (83%) | 586 (80%) |
| Positive | 313 (18%) | 73 (17%) | 145 (20%) |
| **Has ERBB2 alteration** | | | |
| Negative | 1,493 (87%) | 362 (85%) | 636 (87%) |
| Positive | 224 (13%) | 63 (15%) | 95 (13%) |
| **Has RB1 alteration** | | | |
| Negative | 1,369 (80%) | 333 (78%) | 597 (82%) |
| Positive | 348 (20%) | 92 (22%) | 134 (18%) |
| **Has MDM2 alteration** | | | |
| Negative | 1,599 (93%) | 391 (92%) | 663 (91%) |
| Positive | 118 (6.9%) | 34 (8.0%) | 68 (9.3%) |

[1]n (%)